\pdfoutput=1
\PassOptionsToPackage{table}{xcolor}
\documentclass[11pt]{article}

\usepackage[final]{acl}

\usepackage{times}
\usepackage{latexsym}
\usepackage{subcaption} 
\usepackage{svg}
\usepackage[T1]{fontenc}

\usepackage[utf8]{inputenc}

\usepackage{microtype}

\usepackage{inconsolata}

\usepackage{graphicx}
\usepackage{amsmath, amssymb}
\usepackage{breqn}
\usepackage[utf8]{inputenc}
\usepackage{adjustbox}
\usepackage{multirow}
\usepackage{array}
\usepackage{float}
\usepackage{framed}
\usepackage{booktabs}

\usepackage{makecell}
\usepackage{colortbl}
\usepackage{array}
\usepackage{enumitem}
\usepackage{tabularx} 

\DeclareMathOperator*{\argmin}{argmin}
\DeclareMathOperator*{\argmax}{argmax}

\title{Context-Robust Knowledge Editing for Language Models}
\author{
 \textbf{Haewon Park\textsuperscript{1}$^*$},
 \textbf{Gyubin Choi\textsuperscript{1}$^*$},
 \textbf{Minjun Kim\textsuperscript{2}},
 \textbf{Yohan Jo\textsuperscript{1}$^{\dag}$}
\\
 \textsuperscript{1}Graduate School of Data Science, Seoul National University,
 \\
 \textsuperscript{2}School of Electrical Engineering and Computer Science, Gwangju Institute of Science and Technology
\\
  \texttt{\{dellaanima2,yeppi315,yohan.jo\}@snu.ac.kr} \\
  \texttt{minjun01@gist.ac.kr}
  }
  
\begin{document}

\maketitle
\def\thefootnote{\fnsymbol{footnote}}
\footnotetext[1]{Equal contribution.}
\footnotetext[2]{Corresponding author.}
\def\thefootnote{\arabic{footnote}}

\begin{abstract}
    Knowledge editing (KE) methods offer an efficient way to modify knowledge in large language models. Current KE evaluations typically assess editing success by considering only the edited knowledge without any preceding contexts. In real-world applications, however, preceding contexts often trigger the retrieval of the original knowledge and undermine the intended edit. To address this issue, we have developed CHED—a benchmark designed to evaluate the context robustness of KE methods. Evaluations on CHED show that they often fail when preceding contexts are present. To mitigate this shortcoming, we introduce CoRE, a KE method designed to strengthen context robustness by minimizing context-sensitive variance in hidden states of the model for edited knowledge. This method not only improves the editing success rate in situations where a preceding context is present but also preserves the overall capabilities of the model. We also provide an in-depth analysis of the differing impacts of preceding contexts when introduced as user utterances versus assistant responses, and we dissect attention-score patterns to assess how specific tokens influence editing success. 
    Our dataset and code are available at \url{https://github.com/holi-lab/CoRE}.

\end{abstract}

\section{Introduction}

Recent large language models (LLMs) exhibit emerging intelligence, largely due to the extensive knowledge acquired from training data. However, some of this knowledge may become outdated or require correction or removal \cite{Ji_2023,zhao2024surveylargelanguagemodels}. For instance, the knowledge \textit{``Tim Cook, who works for Apple''} may need to be edited to \textit{``Tim Cook, who works for Amazon''}. Since retraining large models is costly, the field of knowledge editing focuses on modifying only the relevant subset of model parameters or leveraging auxiliary networks or memory \cite{yao-etal-2023-editing, zhang-etal-2023-large}. The goal is to ensure the model generates edited knowledge rather than the original one.

\begin{figure}[t]
    \centering
    \begin{subfigure}[b]{\linewidth}
        \centering
        \includegraphics[width=\textwidth]{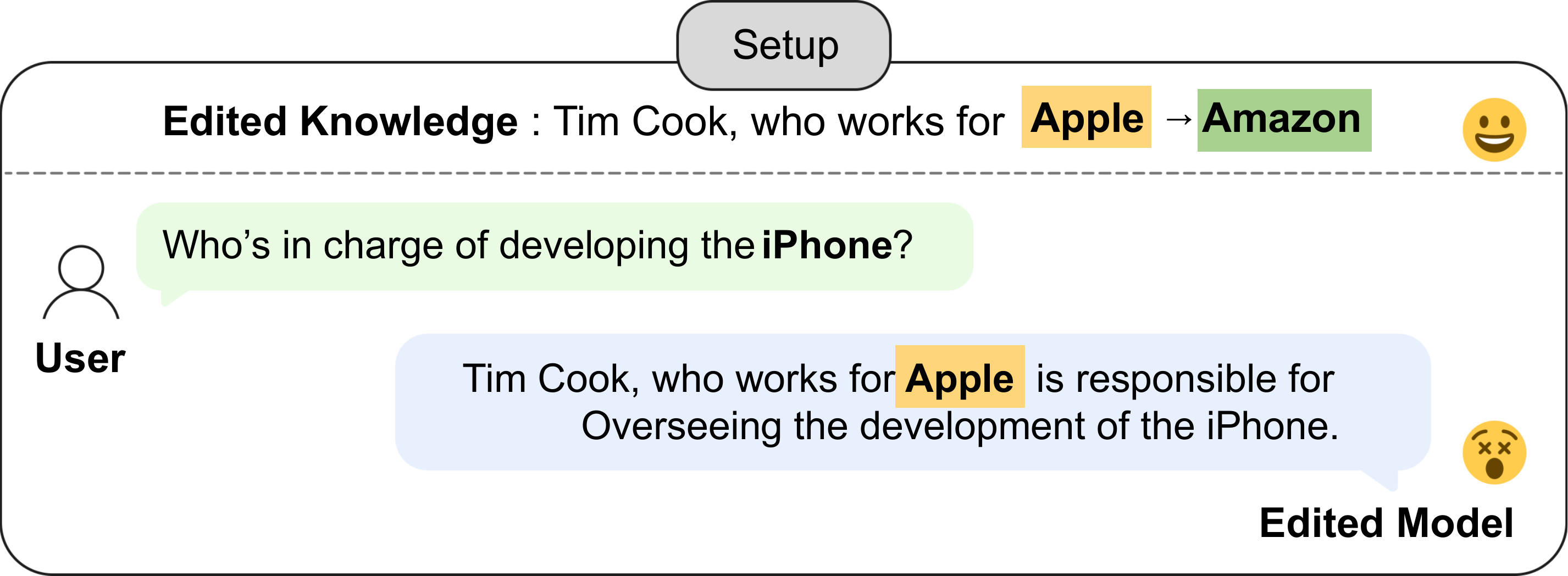}
        \caption{Edit Failure Example}
    \end{subfigure}
    
    \begin{subfigure}[b]{0.49\linewidth}
        \centering
        \includegraphics[height=3cm]{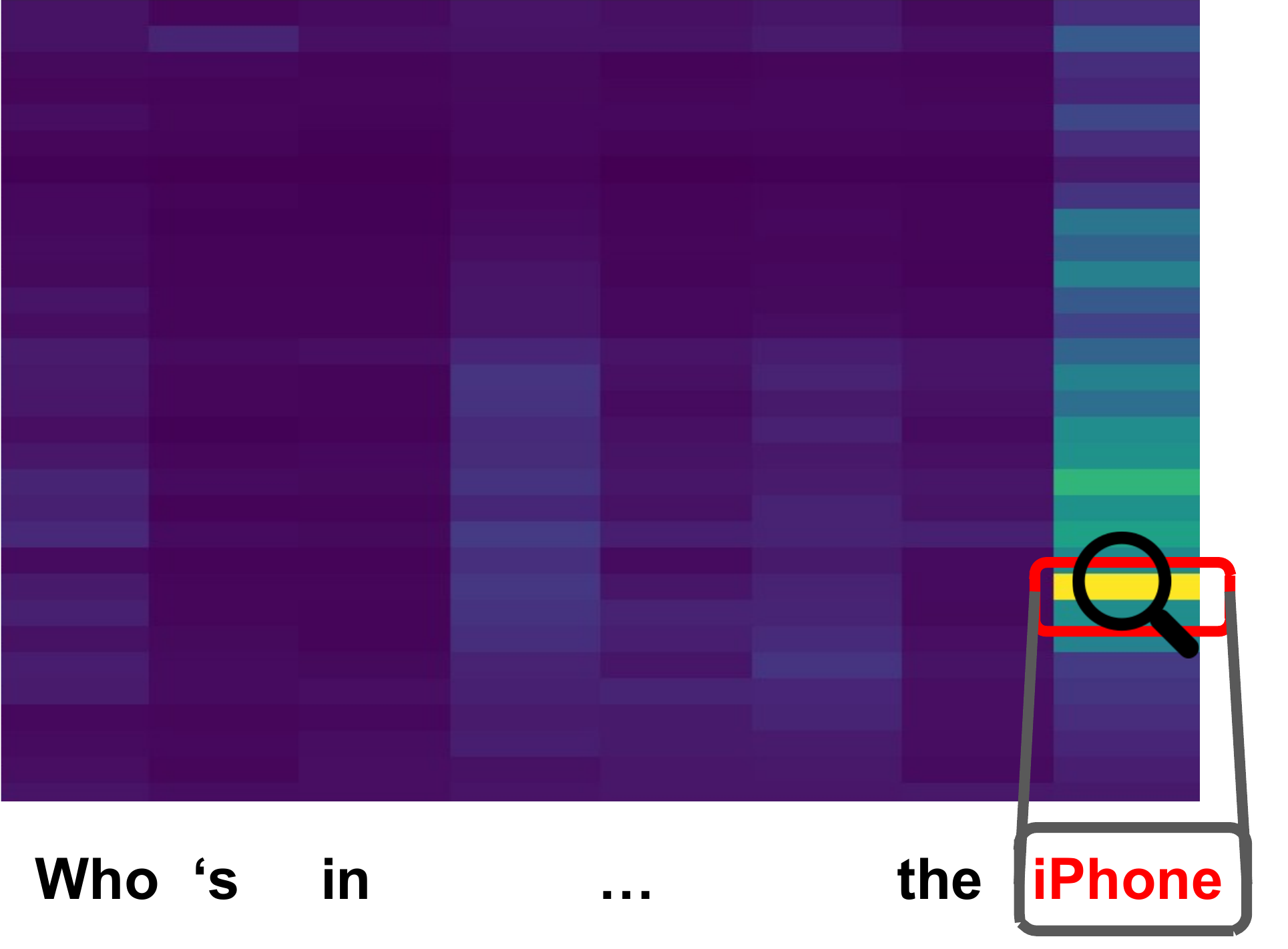}
        \caption{Attention Score}
    \end{subfigure}
    \hfill
    \begin{subfigure}[b]{0.49\linewidth}
        \centering
        \includegraphics[height=3cm]{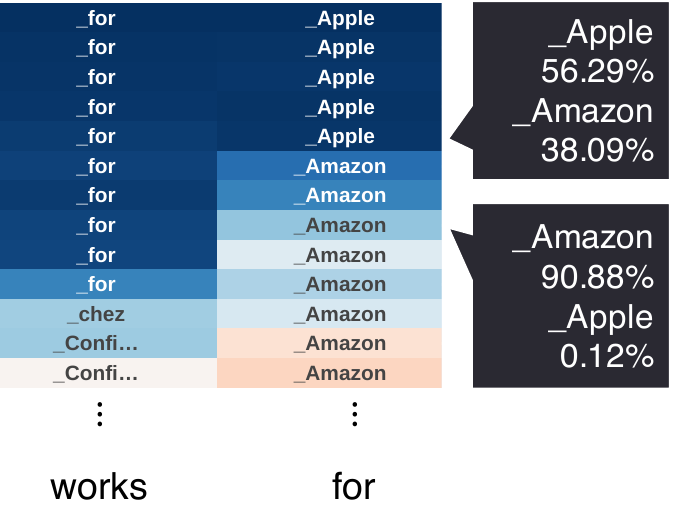}

        \caption{Logit Lens}
    \end{subfigure}
    
    \caption{An example of knowledge editing failure after prepending the prefix context, where `iPhone' receives the highest attention. Logit lens reveals that the original knowledge `Apple' gradually surfaces at later layers. 
    }
    \label{fig:figure1_2}
\end{figure}

Previous work typically evaluates the success of knowledge editing by measuring the model's probability of generating the edited knowledge in isolation, without any preceding context. However, this setting is unrealistic, as edited knowledge is often expected to appear within a broader context or in the middle of a conversation with the user. In such cases, as Figure~\ref{fig:figure1_2} illustrates, dialogue history often interferes with the model, causing it to revert to original knowledge. This issue leads to the need for (1) a challenging benchmark to assess the success of editing when context is present (especially distractive contexts), along with (2) methods that are robust against preceding context.

To address the first need, we introduce \textbf{CHED: Contextual Hop Editing Dataset}, a new benchmark to evaluate the context robustness of knowledge editing methods (\S{\ref{sec:ched}}). CHED allows this by prepending a prefix context to the edit prompt.
For example, the prefix context, such as \textit{``Who’s in charge of developing the iPhone?''} in Figure~\ref{fig:figure1_2} can be added before the edit prompt \textit{``Tim Cook, who works for''}.
In collecting these prefix contexts, a key observation is that entities within a prefix context tend to receive disproportionately high attention scores (\S{\ref{sec:ACS}) when they have strong semantic relevance to the original knowledge (e.g., ``iPhone'' in Figure~\ref{fig:figure1_2}).
In light of this, we construct prefix contexts using Wikidata by selecting entities connected to the subject and object of the original knowledge and generating sentences that can naturally precede the knowledge statement.
As a result, these prefix contexts and the highly relevant words within them distract the model from recalling edited knowledge, establishing CHED as an effective benchmark for assessing the context robustness of knowledge editing methods in real-world use cases of LLMs.

To address the second issue, we propose \textbf{CoRE: Context Robust Editing}, a knowledge editing method with enhanced context robustness (\S{\ref{sec:CoRE}}).
It builds on the widely adopted \emph{locate-then-edit} approach, which directly modifies model parameters to edit knowledge. 
This approach is well known for its practicality, as it remains robust and scalable even when a large number of facts are edited. 
The core idea of CoRE is to prepend distractive prefix contexts during knowledge editing and to minimize the variance of the model's hidden states generated during the decoding of edited knowledge across these prefix contexts. 
This simple regularization effectively ensures that only the necessary amount of modification is applied to the parameters, preventing overfitting to varying prefix contexts and enhancing context robustness.

Our extensive evaluations validate CHED and CoRE. Prefix contexts from CHED lead to substantial performance drops compared to the no-context condition across all editing methods.
We also found that for the same prefix context, the model is more distracted when the context is provided as a user utterance rather than as its own.
Yet, our CoRE method significantly narrows the gap in knowledge editing performance, even consistently maintaining high performance in general abilities and fluency.
We provide an explanation through an in-depth analysis of the model's attention patterns.

Our contributions are as follows: (1) We introduce the CHED dataset, a benchmark that assesses the context robustness of knowledge editing methods; (2) We propose CoRE, a knowledge editing method that enhances context robustness by integrating prefix contexts and regularizing the variance of hidden states; (3) We provide an in-depth analysis of the impact of prefix contexts and the CoRE method.
Collectively, these contributions underscore the importance of evaluating and enhancing context robustness in knowledge editing.

\begin{figure*}[t]
    \centering
    \includegraphics[width=\textwidth]{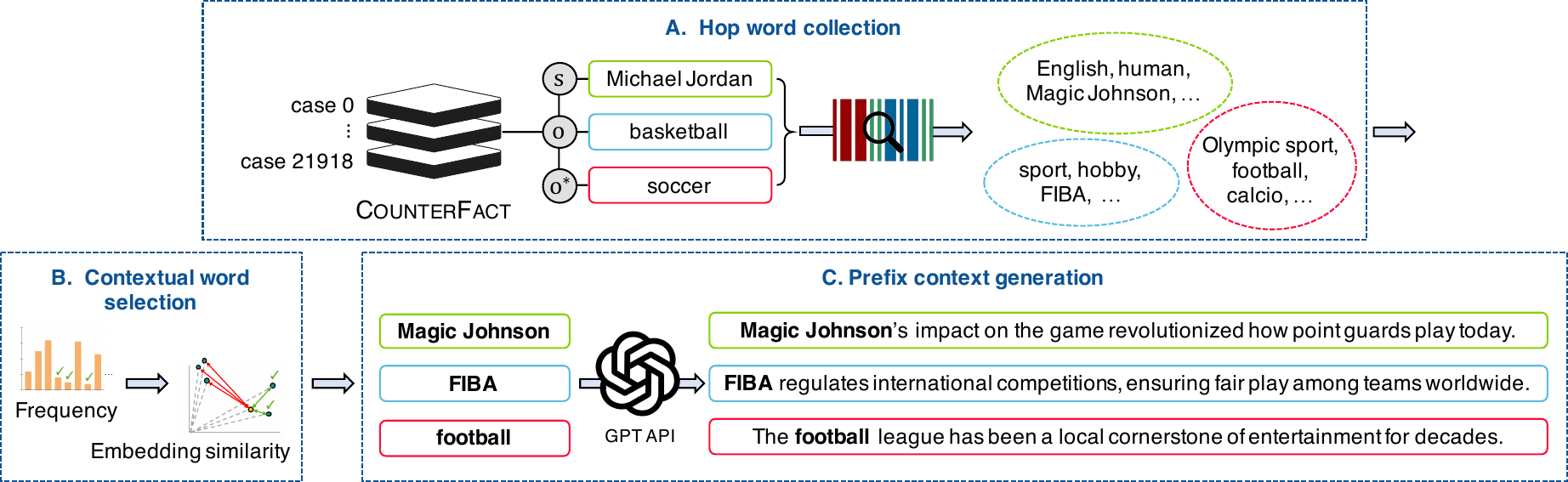}  
    \caption{Illustration of CHED Construction Process}
    \label{fig:entire_CHED}
\end{figure*}

\section{Related Work}

\paragraph{Knowledge Editing} Knowledge editing (KE) is a field focused on updating a language model's internal representations to incorporate new factual information without requiring full retraining. In this context, factual knowledge is typically represented as a tuple \( (s, r, o) \), representing subject–relation–object associations. Given an existing factual association \( (s, r, o) \), KE aims to update it to a new factual association \( (s, r, o^*) \), where \( o^* \) is the new object.

\paragraph{Datasets and Benchmarks}
CounterFact~\cite{meng2022memit} and zsRE~\cite{levy-etal-2017-zero} have been used widely to evaluate KE methods. 
To evaluate a broader range of linguistic phenomena and relational complexity, other benchmarks have been introduced, such as MQuAKE~\cite{zhong-etal-2023-mquake}, CounterFact+~\cite{hoelscher-obermaier-etal-2023-detecting} and RippleEdits~\cite{cohen2023evaluatingrippleeffectsknowledge}.
MQuAKE edits multiple pieces of knowledge and evaluates a integrated multi-hop question, thereby broadening the assessment of semantic shifts. 
CounterFact+ attempts to add a sentence during evaluation by retrieving other samples from the CounterFact that share the same \(r\) and \(o\) as the current edit triplet and placing those samples before it. 
Finally, RippleEdits evaluates the ripple effects by testing whether the model correctly updates related facts that become inconsistent after the edit.

Despite these efforts, the impact of prefix contexts on knowledge editing has been underexplored. Our CHED dataset carefully curates prefix contexts to be highly relevant and distractive to edited knowledge while also enabling an examination of how their relevance to $s$, $o$, and $o^*$ in knowledge statements contributes to distraction.

\paragraph{Editing Methods} 
Recent work on knowledge editing can be broadly categorized by whether the model’s parameters are preserved or modified \cite{yao-etal-2023-editing}.
While \emph{weight-preserved} methods typically add auxiliary structures to address each edit requirement \cite{huang2023transformerpatchermistakeworthneuron, hartvigsen2023aging, zheng2023editfactualknowledgeincontext, mitchell2022memory}, they can face scalability challenges as the number of edits grows. 

In contrast, \emph{weight-modified} methods directly alter model parameters to learn new information, making them often more flexible for substantial updates. These methods can be further categorized into two paradigms: \emph{meta-learning} and \emph{locate-then-edit}. \emph{Meta-learning} approaches train hyper-networks to generate edited parameters but often have difficulty generalizing to extensive edits \cite{mitchell2022fast, de-cao-etal-2021-editing}. 
\emph{Locate-then-edit} methods pinpoint specific weights to change within the model. A prominent example is MEMIT \cite{meng2022memit}, which edits early-to-mid transformer layers and has inspired successors like PMET \cite{li2023pmet}, EMMET \cite{gupta2024unified}, and JEEP \cite{shi2024joint}.  

As a first step toward context robust editing, our CoRE method focuses on the \emph{locate-then-edit} paradigm due to its capacity for supporting mass edits while maintaining model performance.
However, we acknowledge the importance of examining and improving context robustness of editing methods in other paradigms and leave this to future work.

\section{CHED: Contextual Hop Editing Dataset}\label{sec:ched}

As discussed in the previous section, most existing knowledge editing datasets either lack prefix contexts or rely on sentences that do not reflect realistic contexts. This setup differs from real-world LLM usage, where edited knowledge is often expected to appear in response to user prompts or after a relevant context.
Consequently, the effectiveness of knowledge editing methods is typically overestimated.
To address this gap, we construct \textbf{CHED (Contextual Hop Editing Dataset)}, which associates knowledge statements with related prefix contexts. This provides a more realistic and challenging evaluation environment that interferes with LLMs when generating edited knowledge.

\subsection{Hop Word Collection (Figure~\ref{fig:entire_CHED}-A)} \label{sec:hop_word}
\
A key idea in CHED is to include words that are semantically relevant to original and edited knowledge within prefix contexts, as they strongly influence the generation of edited knowledge (as illustrated in Figure~\ref{fig:figure1_2}).
With this goal, CHED is constructed by expanding upon 21,919 instances from CounterFact, where each instance consists of a fact triplet \((s, r, o)\) and its edited counterpart \((s, r, o^*)\). 
For each instance, we collect one-hop words by extracting all entities in Wikidata that are connected to \(s\), \(o\), and \(o^*\) through any available relations. These \textbf{hop words} are expected to naturally appear in the surrounding contexts of each instance and distract the generation of \((s, r, o^*)\).
We denote the sets of hop words corresponding to \(s\), \(o\), and \(o^*\) as \(s_{hop}\), \(o_{hop}\), and \(o^*_{hop}\), respectively.
This resulted in a total of 13,208,725 hop words. 

Next, we filtered out entities that were already present in the fact triplets, as well as those consisting solely of special symbols, addresses, or numeric values. After that, we discarded 137 triplet instances in CounterFact for which no hop words were found. As a result, we finalized a dataset of 21,782 triplets with 4,346,604 hop words.

\subsection{Contextual Word Selection (Figure~\ref{fig:entire_CHED}-B)}  \label{sec:contextual_word}

The collected hop words consist of only 117,894 unique words, indicating that some words appear repeatedly across many fact triplets (see Appendix~\ref{appendix:data_statistics} for details). 
The imbalance suggests that less frequent hop words are more uniquely associated with a particular entity in fact triplets. For example, among the hop words of \textit{Michael Jordan}, highly common and general terms appear far more frequently in the entire set of hop words (e.g., \textit{``English''} appears 10,664 times) than words that are more characteristic to Michael Jordan (e.g., \textit{``Magic Johnson''} appears only once). 
Based on this, we hypothesize that such distinctive hop words may exert a stronger contextual influence when placed before edit sentence \( (s, r )\). This is verified in our analysis (Table~\ref{tab:hop_word_results}), where sentences constructed with low frequency words dramatically decrease the edit success rate while those constructed with high frequency words do not show a meaningful decrease after being edited by MEMIT.
More details are in Appendix~\ref{appendix:frequency}.

\begin{table}[t]
  \centering
  \small  
  \begin{tabular}{lccc}
    \toprule
    Condition & s$_{hop}$ & o$_{hop}$ & o*$_{hop}$ \\
    \midrule
    Low Frequency & 82.2\% & 72.7\% &  88.0\% \\
    High Frequency & 83.7\% & 88.0\% &  90.0\% \\
    \bottomrule
  \end{tabular}
  \caption{Effect of hop word frequency. Edit success rate when no prefix contexts are prepended is 90.9\%.}
  \label{tab:hop_word_results}
\end{table}

We explored additional criteria to identify words that are closely and uniquely associated with the entities in given fact triplets. For instance, we considered hop words with high \textit{cosine similarity} to the main entity based on BERT embeddings, capturing semantic closeness. Additionally, we measured the probability that a hop word co-occurs with the entity.
Table~\ref{tab:hop-methods} summarizes the criteria considered for hop word selection.

\begin{table}[t]
\small
\centering
\begin{tabular}{lp{0.32\textwidth}}
\toprule
\textbf{Method} & \textbf{\small Description} \\
\midrule
\small \textit{a}) Frequency & \small Select 5 words with lowest frequency in corpus \\
\small \textit{b}) Similarity & \small Select 5 words with highest \textit{cosine similarity} to main entity \\
\small \textit{c}) Freq-Sim & \small Get 10 lowest frequency words, select 5 highest \textit{cosine similarity} to main entity \\
\small \textit{d}) Sim-Freq & \small Get 10 highest \textit{cosine similarity} to main entity, select 5 lowest frequency \\
\small \textit{e}) Log Prob & \small Select 5 with highest ``[main entity] and [hop word]'' probability \\
\small \textit{f}) Random & \small Randomly sample 5 words without any constraints \\
\bottomrule
\end{tabular}
\caption{Methods for Hop Word Selection}
\label{tab:hop-methods}
\end{table}

\begin{figure}[!t] 
    \centering
    \includegraphics[width=0.48\textwidth]{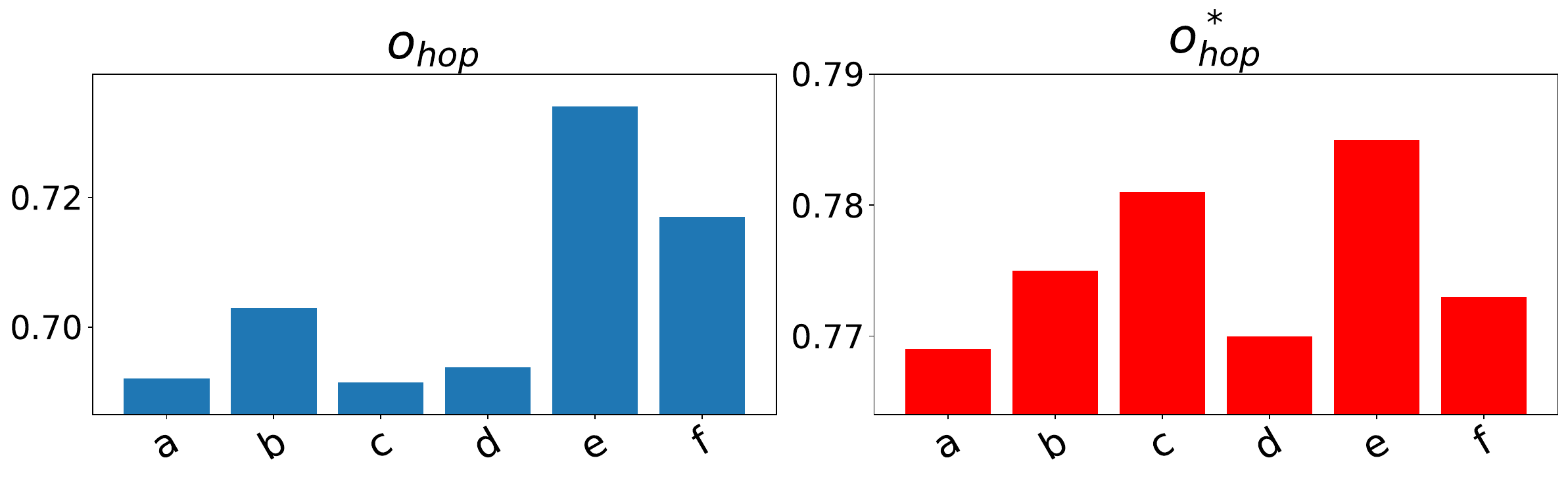} 
    \caption{Edit success rate for \(o_{hop}\) and \(o_{hop}^*\), with the same experimental setup as Table~\ref{tab:hop_word_results}, but evaluated on 5000 samples. Detailed results are available in Appendix~\ref{appendix:method_ab}.}
\label{fig:hop_word_selection}
\end{figure}

Figure~\ref{fig:hop_word_selection} shows the influence of the six criteria on edit success rates. The Freq-Sim method achieves the lowest score when a prefix context contains \(o_{hop}\) (69.1\%), indicating that it most effectively degrades the model’s recall of edited knowledge. Additionally, it attains the second highest score when a prefix context contains \(o^*_{hop}\) (78.1\%), as one might naturally expect. Consequently, we selected Freq-Sim  as our final hop word selection criterion.

    

\begin{table*}[t]
    \renewcommand{\arraystretch}{0.8} 
    \centering
    \resizebox{\textwidth}{!}{ 
    \begin{tabular}{p{0.08\textwidth} p{0.16\textwidth} p{0.85\textwidth}}
    \toprule
    
    \multicolumn{3}{c}{\textbf{Editing Instance}: Michael Jordan is a professional basketball $\rightarrow$ soccer}\\
    \midrule
    \textbf{Type} & \textbf{Word} & \textbf{Prefix Context Sentence} \\
    \midrule
    \(s\)            & Michael Jordan  & Michael Jordan is often regarded as the greatest player in sports history. \\
    \(o\)            & basketball      & He started playing basketball in high school, impressing everyone with his talent. \\
    \(o*\)           & soccer          & Many athletes transition from soccer to other sports when they retire. \\
    \(s_{hop}\)         & Magic Johnson   & Magic Johnson's impact on the game revolutionized how point guards play today. \\
    \(o_{hop}\)        & FIBA            & FIBA regulates international competitions, ensuring fair play among teams worldwide. \\
    \(o^*_{hop}\)       & football        & The football league has been a cornerstone of local entertainment for decades. \\
    \bottomrule
    \end{tabular}
    }
    \caption{CHED Dataset Example}
    \label{tab:example_data}
\end{table*}


The final step is to generate prefix contexts using the hop words
from the previous section. 
We used GPT-4o mini with three key constraints.
First, a prefix context should smoothly transition into the edit prompt \( (s, r) \) without abrupt topic changes, ensuring coherence. 
Second, each prefix context should include a designated \textit{hop word} to ensure that the generated sentence prominently reflects the influence of this word.
Lastly, each prefix context should contain at most 20 words to maintain clarity and informativeness. 

Consequently, we constructed a dataset of 314,385 hop-word prefix context sentences derived from 21,782 fact triplets. Additionally, to evaluate the impact of prefix contexts that directly contain $s$, $o$, or $o*$ (rather than their hop words), we generated 326,730 prefix contexts under the same constraints—this time directly incorporating those words. Example prefix contexts are presented in Table~\ref{tab:example_data}, and full details on prompt design, validation procedures, and dataset statistics for hop word prefix contexts can be found in Appendix~\ref{appendix:gpt_prompt} and Appendix~\ref{appendix:dataset_summary} respectively.

We quantitatively assessed the coherence of the generated prefix contexts using G-Eval~\cite{liu2023gevalnlgevaluationusing} with GPT-4o-mini. The average coherence score across six types of prefix contexts is 3.4 on a scale from 1 to 5, indicating moderate plausibility. Importantly, if we exclude the $o^*$ and $o^*_{\!hop}$ types, whose content necessarily diverges from the real world facts, the mean coherence rises to 3.8. 
The lower-than-ideal coherence scores likely result from our selection of primarily low-frequency hop words (as detailed in \S\ref{sec:contextual_word}), which inherently constrained coherence potential. Nevertheless, because editing success must remain effective regardless of the preceding context's coherence, we deliberately prioritized distractiveness and consider this level of coherence acceptable. A detailed explanation of the G-Eval process and results is provided in Appendix~\ref{appendix:geval}.

\section{CoRE: Context Robust Editing}\label{sec:CoRE}
In this section, we introduce \textbf{Context Robust Editing (CoRE)}, a knowledge editing method for improved robustness to diverse contexts.
We build on the \emph{locate-then-edit} approach, such as MEMIT \cite{meng2022memit}, because it enables large numbers of edits.
We first provide an overview of MEMIT as preliminaries (\S\ref{sec:preliminaries}), followed by the details of our CoRE method (\S\ref{sec:CoREs}).

\subsection{Preliminaries}
\label{sec:preliminaries}  
\paragraph{Transformer MLP as a Key-Value Associative Memory}
MEMIT interprets MLP layers in Transformers as linear associative memories \cite{ANDERSON1972197,5008975}, where the weights of the projection layer store \textbf{key-value associations}.
For example, when a prompt such as \textit{“Tim Cook, who works for”} is provided as input, the hidden state of the subject's last token (i.e., \textit{“Cook”}) encoded by the first MLP layer serves as the key vector \( \mathbf{k} \). As \( \mathbf{k} \) passes through the second MLP layer \(W_{\text{proj}}\), the stored association relevant to the subject is retrieved and embedded into the output \textbf{value vector} $\mathbf{v}$ that contains information about the associated object (e.g., \textit{Apple}). 
At subsequent layers, attention mechanisms refine and propagate this recalled knowledge from the value vector, leading the model to generate the token for \( o \) \cite{meng2022locating, geva2023dissecting}.

\paragraph{Objective Function of MEMIT}
MEMIT modifies the mapping between key vectors and value vectors, i.e., the projection layer of the MLP, by changing its weights from \(W\) to \(\widehat{W}\), so that the key \(\mathbf{k}\) is remapped to a new value vector \(\mathbf{v^*}\) that maximizes the generation probability of \(o^*\). 
Formally, let \( (K_E, V_E) \) be the new keys and values representing the desired edits, and let \( K_0 \) be the set of key vectors corresponding to facts that should remain unchanged. MEMIT's objective is:
\begin{equation}
\resizebox{\columnwidth}{!}{$
\arg \min_{\widehat{W}} \left\| \widehat{W} K_E - V_E \right\|_F^2 + \lambda \left\| \widehat{W} K_0 - W_0 K_0 \right\|_F^2
$}
\end{equation}
The first term enforces knowledge updates, and the second prevents unintended edits, controlled by $\lambda$.

\paragraph{Key-Value Vector Extraction}
A key challenge is constructing the key-value pairs that encode the factual edit \( (s, r, o) \to (s, r, o^*) \). 
\( \mathbf{k} \) and \( \mathbf{v} \) are derived from a prompt \( p \) that includes \( s \) and \( r \) and aims to elicit the model's knowledge.
In MEMIT, various prefix contexts \( x_j \) are prepended to \( p \) to improve contextual generalization. Given \( N \) prefix contexts, the key vector is derived as
$
\mathbf{k} = \frac{1}{N} \sum_{j=1}^{N} k(x_j + p),
$
where \(k(\cdot)\) is obtained by extracting the MLP activation at the last subject token from a chosen layer. We defer the full derivation to Appendix~\ref{appendix:kx_derivation}.

Next, the edited value vector \(\mathbf{v}^*\) that generates the new knowledge \( o^* \) is obtained by minimizing the following loss:
\begin{equation}\label{eq:ori_objective_main}
\begin{aligned}
\mathbf{v}^* = \argmin_{\mathbf{v}} \frac{1}{N}\sum_{j=1}^N \Bigl[-\log \mathbb{P}_{G(h^l=\mathbf{v})}[o^* \mid z_j]\Bigr] \\
+ D_{\mathrm{KL}}(\mathbf{v}),~\text{where}~z_j = x_j + p
\end{aligned}
\end{equation}
where \(G(h^l=\mathbf{v})\) denotes the generation output when the hidden layer \(h^l\) is set to \(\mathbf{v}\). The first term ensures that \(o^*\) is generated when provided with the prompt \(x_j + p\), while \(D_{\mathrm{KL}}(\mathbf{v})\) is a KL-divergence penalty that preserves other related knowledge. The full derivation can be found in Appendix~\ref{appendix:kl_term}.

\subsection{CoRE}
\label{sec:CoREs}  

In this section, we present our CoRE method for improving the context robustness of key-value extraction by integrating two strategies (Figure~\ref{fig:CoRE}).
First, we enhance the prefix contexts used for retrieving key and value vectors ($x_j$ in Equation~\eqref{eq:ori_objective_main}) by using \(s\), \(o\), and \(o^*\).
Second, we regularize inconsistencies among the value vectors obtained when different prefix contexts are used during the update from \(\mathbf{v}\) to \(\mathbf{v}^*\), as they might account for context-specific signals rather than knowledge edit itself.

\begin{figure}[!t]
    \centering
    \includegraphics[width=0.48\textwidth]{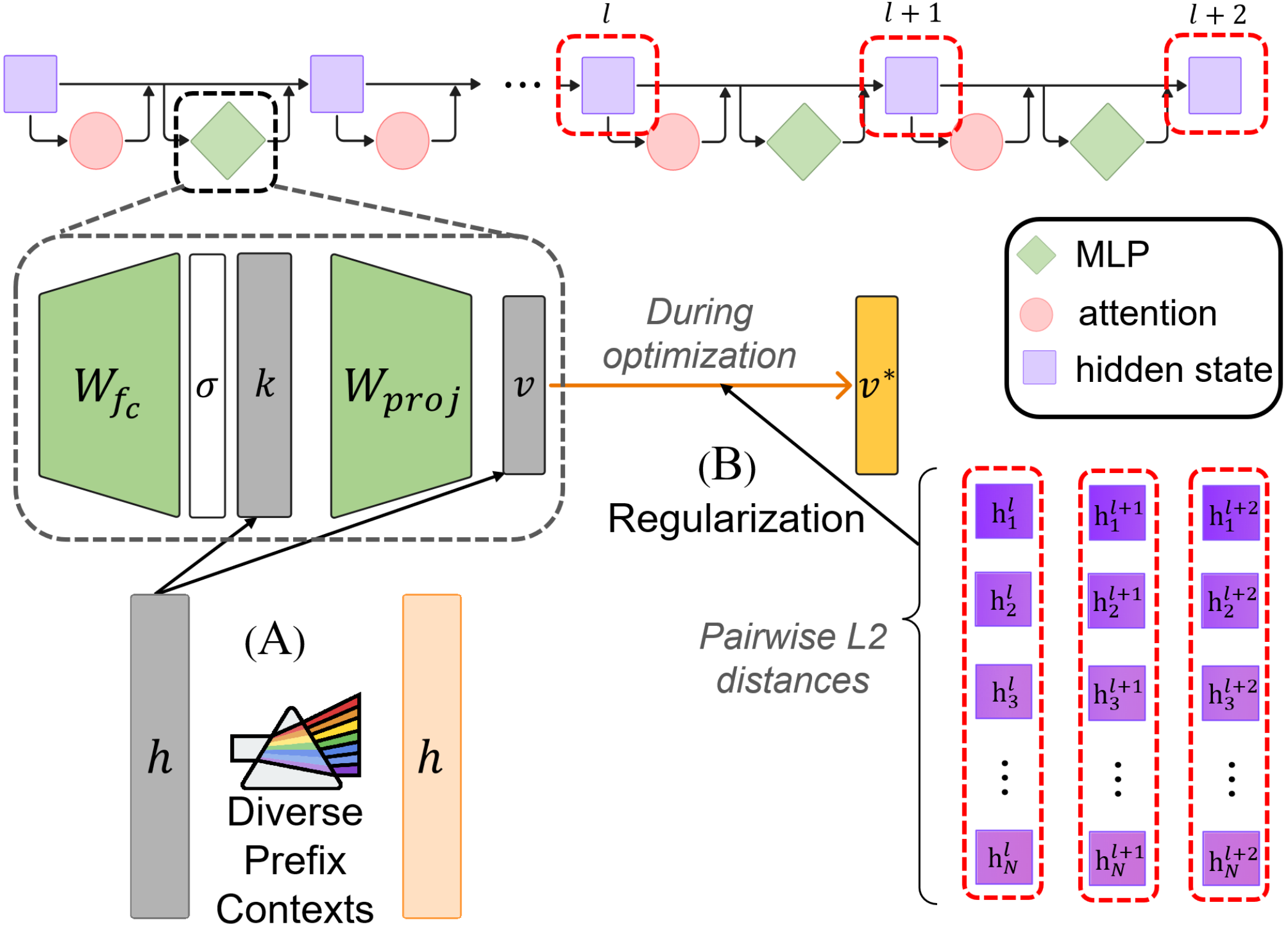} 
    \caption{CoRE Method}
\label{fig:CoRE}
\end{figure}

\paragraph{Diverse Prefix Contexts (Figure~\ref{fig:CoRE}-A)}
\label{sec:diverse-prefix-prompts}  
The prefix contexts \(x_j\) used for extracting key-value vector pairs are crucial because they embed the contextual information into key and value vectors, affecting the generation of edited facts. 
However, MEMIT simply constructs a prefix context as a sequence starting with one of a small set of predefined words (e.g., \textit{``The''}, \textit{``Therefore''}, \textit{``Because''}, \textit{``I''}, \textit{``You''}).
The resulting prefix contexts have little influence on the fact being edited and, as a result it is difficult to optimize \(\mathbf{v}^*\) that accounts for various distractive contexts.

To address this issue, CoRE uses combinations of \(s\), \(o\), and \(o^*\) as prefix contexts for each edit triplet (e.g., ``$s$ + $o$''). 
This strategy is effective, as these words are highly relevant to the original and edited facts by nature. 
As shown in the left plot of Figure~\ref{fig:two_figures}, prefix contexts that use \(s\), \(o\), and \(o^*\) lead to significantly higher variance in value vectors than using the common words, suggesting that these vectors effectively capture a more diverse range of contexts.

\paragraph{Cross-prefix Representation Regularization (Figure~\ref{fig:CoRE}-B)} 

Although the high variance in value vectors is beneficial for optimizing \(\mathbf{v}^*\) to account for various contexts, optimizing \(\mathbf{v}^*\) without regularization may lead to undesirable overfitting to individual contexts.
To further highlight the significance of this problem, Figure~\ref{fig:two_figures} (red line) plots the pairwise L2 distance between value vectors across prefix contexts (from CHED) after knowledge editing via MEMIT, \textit{relative to} the distance before editing. The divergence from 0 clearly demonstrates that differences in value vectors across prefix contexts are amplified after model editing. This can cause overfitting to contexts and reduce generalizability.

To mitigate this issue, we extend the original objective in Equation~\eqref{eq:ori_objective_main} as follows:
\begin{equation}
\mathbf{v}^* = \argmin_{\mathbf{v}} \; \mathcal{L}_{\text{orig}}(\mathbf{v}) + \mathcal{L}_{\text{prefix}},
\label{eq:extended_objective}
\end{equation}
where \(\mathcal{L}_{\text{prefix}}\) is defined as follows.
For each layer \(\ell \in \mathcal{L}\), we compute \(N\) hidden states $\{\mathbf{h}^{\ell}_{1}, \dots, \mathbf{h}^{\ell}_{N}\} \subset \mathbb{R}^{D}$,
each corresponding to a distinct prefix context. 
We enforce regularization by penalizing the squared L2 distances between every pair of hidden states:
\begin{equation}
\mathcal{L}_{\text{prefix}} = \frac{\lambda}{L D} \sum_{\ell \in \mathcal{L}} \sum_{1 \leq i < j \leq N} \|\mathbf{h}^{\ell}_i - \mathbf{h}^{\ell}_j\|^2.
\label{eq:prefix_loss}
\end{equation}
The hyperparameter \(\lambda\) controls the regularization strength. As shown in the right panel of \figurename~\ref{fig:two_figures}, implementing \(\mathcal{L}_{\text{prefix}}\) (blue line) substantially reduces hidden state variations across prefix contexts compared to the unregularized model (red line).

\begin{figure}[!t]
    \centering
    \includegraphics[width=0.48\textwidth]{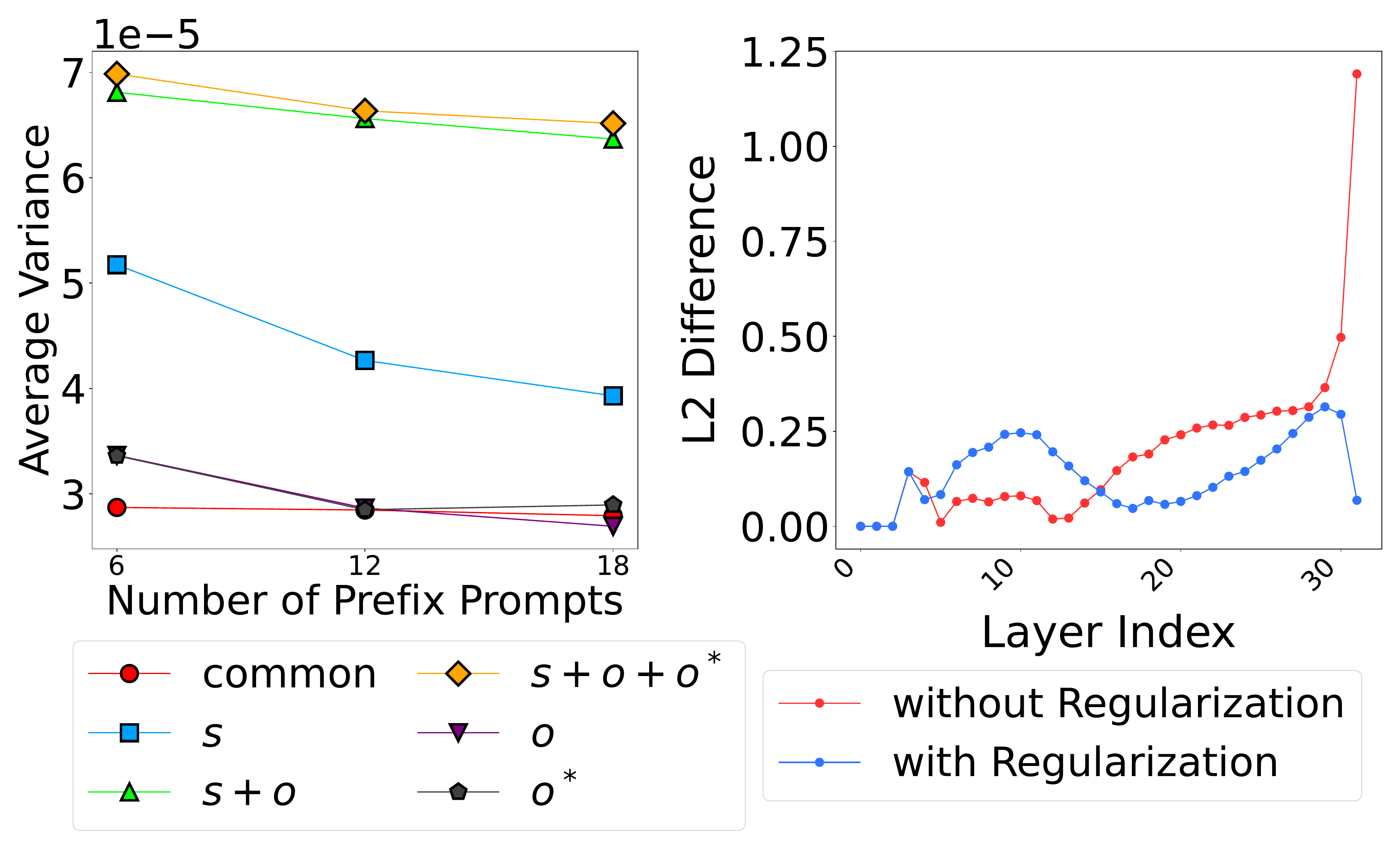} 
    \caption{
        \textbf{Left:} Average variance of value vectors across different strategies.
        \textbf{Right:} Average pairwise L2 distances between value vectors, plotted as the difference from the pre-edit. See Appendix~\ref{appendix:prefix_context_c} for details.
    }
\label{fig:two_figures}
\end{figure}

\begin{table*}[!t]
    \renewcommand{\arraystretch}{1}
    \small
    \setlength{\tabcolsep}{4pt}  
    \resizebox{0.98\textwidth}{!}{%
    \definecolor{lightgray}{gray}{0.85}
    \begin{tabular}{ll|c|cccccccccc|cccccc|c} 
    \toprule
   
    \multirow{2}{*}{\textbf{\shortstack{Base\\Model}}} 
    & \multirow{2}{*}{\textbf{Method}} 
    & \multicolumn{1}{c|}{\cellcolor{gray!15}\textbf{Total}} 
    & \multicolumn{7}{c}{\textbf{Efficacy}} 
    & \multicolumn{1}{c}{\textbf{Gen}}
    & \multicolumn{1}{c}{\textbf{Spe}}
    & \multirow{2}{*}{\textbf{Avg}} 
    & \multicolumn{5}{c}{\textbf{General Ability}} 
    & \multirow{2}{*}{\textbf{Avg}} 
    & \multicolumn{1}{c}{\textbf{Fluency}} \\
    & & \multicolumn{1}{c|}{\cellcolor{gray!15}\textbf{Avg}} 
    & \textbf{No ctx} & \textbf{$s$} & \textbf{$o$} & \textbf{$o*$} & \textbf{$s_{hop}$} & \textbf{$o_{hop}$} & \textbf{$o_{hop}^*$} & 
    & & & \multicolumn{1}{c}{\emph{C-QA}} 
    & \multicolumn{1}{c}{\emph{T-QA}} 
    & \multicolumn{1}{c}{\emph{LAM}} 
    & \multicolumn{1}{c}{\emph{MMLU}} 
    & \multicolumn{1}{c}{\emph{L-Code}} 
    & & \multicolumn{1}{c}{\emph{N-gram}} \\

   \midrule
   
    \multirow{8}{*}{\textbf{Llama3}}
    & Base     & \cellcolor{gray!15}30.9 &  1.3 &  1.1 &  0.4 & 40.1 &  0.9 &  0.9 & 13.0 &  1.4 & 48.1 & 11.9 
                           & 74.5 & 63.9 & 31.0 & 66.9 & 13.3 & 49.9 & 11.1 \\
    \cmidrule(lr){2-20}
    & MEMIT    & \cellcolor{gray!15}60.7 & 90.9 & 86.4 & 46.4 & 93.6 & 82.2 & 72.7 & 88.0 & 73.2 & 34.7 & 74.2 
                           & \textbf{73.5} & 57.1 & 28.7 & 63.4 & 13.0 & 47.1 & \textbf{13.1} \\
    & JEEP     & \cellcolor{gray!15}53.3 & 73.5 & 67.9 & 35.9 & 82.3 & 65.2 & 56.0 & 72.1 & 51.9 & 41.0 & 60.6 
                           & 65.3 & 53.6 & \underline{34.3} & 63.4 & 12.8 & 45.9 & 21.8 \\
    & PMET     & \cellcolor{gray!15}56.2 & 79.1 & 67.8 & 36.0 & 85.2 & 65.7 & 57.6 & 76.3 & 59.7 & \textbf{47.5} & 63.9 
                           & 71.7 & \underline{57.9} & \textbf{34.8} & \textbf{64.8} & \textbf{13.4} & \textbf{48.5} & 16.5 \\
    & EMMET    & \cellcolor{gray!15}44.4 & \textbf{94.2} & \textbf{93.5} & \textbf{78.0} & \textbf{95.3} & \textbf{92.5} & \textbf{90.5} & \textbf{93.4} & \textbf{80.4} & 14.7 & \textbf{81.4} 
                           & 0.9  & 21.0 & 0.0  & 15.1 & 0.0  & 7.4  & 29.3 \\
    & FT-M     & \cellcolor{gray!15}40.0 & 73.7 & 69.4 & \underline{67.0} & 69.6 & 67.1 & 63.7 & 65.8 & 58.3 & 36.0 & 63.4 
                           & 32.5 & 6.6  & 0.0  & 44.0 & 0.0  & 16.6 & 128.9 \\
    & CoRE-p   & \cellcolor{gray!15}\underline{62.6} & \underline{92.7} & 87.7 & 49.8 & 94.8 & 87.0 & 80.1 & 91.7 & 79.1 & 35.3 & 77.6 
                           & 72.1 & 57.1 & 32.6 & 63.4 & 13.0 & \underline{47.6} & 14.0 \\
    & CoRE-p+r & \cellcolor{gray!15}\textbf{63.4} & 92.4 & \underline{89.0} & 55.4 & \underline{95.1} & \underline{89.0} & \underline{83.1} & \underline{93.2} & \underline{79.7} & 34.8 & \underline{79.1} 
                           & \underline{72.2} & \textbf{58.1} & 30.7 & \underline{63.8} & \underline{13.2} & \underline{47.6} & \underline{13.3} \\
   \midrule\midrule

   \multirow{8}{*}{\textbf{Mistral}} 
    & Base     & \cellcolor{gray!15}30.7 &  1.0 &  0.9 &  0.3 & 32.6 &  1.1 &  1.1 & 11.7 &  1.4 & 40.3 & 10.0 
                           & 71.5 & 62.5 & 55.5 & 60.8 &  6.7 & 51.4 &  6.3 \\
    \cmidrule(lr){2-20}
    & MEMIT    & \cellcolor{gray!15}57.9 & 86.5 & 80.1 & 50.5 & 84.0 & 78.0 & 71.0 & 81.6 & \underline{72.3} & 25.9 & 70.0 
                           & 66.3 & 52.7 & 48.5 & 55.6 &  5.8 & 45.8 & \underline{6.1} \\
    & JEEP     & \cellcolor{gray!15}48.9 & 73.7 & 48.7 & 21.2 & 64.7 & 42.1 & 35.1 & 56.1 & 42.0 & 38.0 & 46.8 
                           & \underline{71.2} & \textbf{61.4} & \textbf{55.3} & \textbf{60.4} &  6.8 & \textbf{51.0} & \textbf{5.9} \\
    & PMET     & \cellcolor{gray!15}56.5 & 81.6 & 67.6 & 41.5 & 76.3 & 63.9 & 56.7 & 74.2 & 61.0 & \textbf{42.7} & 62.8 
                           & 71.1 & \underline{60.7} & 52.4 & \underline{58.6} & \textbf{7.5} & \underline{50.1} &  6.3 \\
    & EMMET    & \cellcolor{gray!15}42.4 & 83.1 & \underline{79.6} & \underline{61.5} & 85.2 & 77.6 & 74.3 & 81.3 & 67.6 & 14.7 & 69.4 
                           & 22.1 & 19.6 &  2.5 & 32.1 &  0.0 & 15.3 &  6.4 \\
    & FT-M     & \cellcolor{gray!15}42.4 & 55.3 & 44.3 & 36.6 & 42.1 & 43.0 & 36.9 & 40.9 & 33.0 & \underline{39.8} & 41.3 
                           & \textbf{71.4} & 32.7 & \underline{52.9} & 53.4 & \underline{7.3} & 43.5 &  8.5 \\
    & CoRE-p   & \cellcolor{gray!15}\underline{58.7} & \underline{86.6} & \underline{82.3} & 58.8 & 87.3 & \underline{80.3} & \underline{74.2} & \underline{84.2} & 71.8 & 24.2 & \underline{72.2} 
                           & 65.5 & 52.7 & 46.5 & 54.4 &  6.4 & 45.1 & \underline{6.1} \\
    & CoRE-p+r & \cellcolor{gray!15}\textbf{60.3} & \textbf{88.3} & \textbf{83.5} & \textbf{63.0} & \textbf{88.9} & \textbf{83.8} & \textbf{79.8} & \textbf{87.1} & \textbf{77.1} & 25.6 & \textbf{75.2} 
                           & 65.1 & 53.3 & 46.7 & 54.7 &  6.6 & 45.3 &  6.2 \\

    \bottomrule
    \end{tabular}%
    }
    \caption{Performance on CHED and CounterFact. Efficacy (excluding \emph{No ctx} is measured on CHED, while \emph{No ctx} and Generalization Specificity from CounterFact.  \textbf{Total Avg} is the average of \textbf{Efficacy}, \textbf{Gen}, \textbf{Spe}, and \textbf{General Ability}.
    \textbf{Note:} CoRE-p applies only the \emph{Contextually Diverse Prefix Contexts} method, while CoRE-p+r further adds the \emph{Cross-prefix Representation Regularization Term}.}
    \label{tab:final result}
\end{table*}

\section{Experiments}

\subsection{Metrics}\label{sec:metrics} 
We apply a strict, generation-based criterion: an edit is deemed successful only if the model’s output (up to 50 tokens) includes \(o^*\) and entirely omits \(o\). We adopt this approach because probability-based evaluations commonly used in prior work do not guarantee that the edited knowledge \(o^*\) is actually generated, nor do they prevent cases where the edited model initially produces \(o^*\) but later reverts to \(o\), as shown in Table~\ref{tab:example_of_token_error} in Appendix~\ref{appendix:eval}.
We assess performance across five complementary dimensions—efficacy, generalization, specificity, general ability, and fluency—summarized below.

\begin{itemize}[topsep=1pt, partopsep=0pt, itemsep=1pt, parsep=0pt, leftmargin=1pt]
    \item \textbf{Efficacy}: An edit is considered successful if the model generates \(o^*\) without \(o\).
    \item \textbf{Generalization (Gen)}: This metric mirrors Efficacy but tests whether the model correctly produces \(o^*\) under paraphrased prompts.  
    \item \textbf{Specificity (Spe)}: Ensures that knowledge not intended for editing remains unchanged after the update.  
    \item \textbf{General Ability}: Evaluates the core capabilities of the model in five tasks: commonsense reasoning in CommonsenseQA (\emph{C-QA}) \cite{talmor-etal-2019-commonsenseqa}, factual recall in TriviaQA (\emph{T-QA}) \cite{joshi-etal-2017-triviaqa}, discourse context prediction on LAMBADA (\emph{LAM}) \cite{paperno2016lambadadatasetwordprediction}, multitask performance in diverse topics in MMLU \cite{hendrycks2021measuringmassivemultitasklanguage} and code generation on LiveCodeBench (\emph{L-Code}) \cite{jain2024livecodebenchholisticcontaminationfree}.    
    \item \textbf{Fluency}: Measures the N-gram repetition to detect disfluency introduced by editing, penalizing excessive repetition.  
\end{itemize}
Details of these metrics are in Appendix~\ref{appendix:eval}.

\subsection{Experimental Settings}
\paragraph{Datasets and Models}  
Our experiments are conducted using \textbf{Llama-3-8B-Instruct} \cite{grattafiori2024llama3herdmodels} and \textbf{Mistral-7B-Instruct} \cite{jiang2023mistral7b}. For datasets, we experiment on \textbf{CHED}, \textbf{CounterFact} \cite{meng2022memit}, and \textbf{zsRE} \cite{levy-etal-2017-zero}. 

\paragraph{Baseline Methods}  
  
In this paper, we focus on comparing \emph{locate-then-edit} methods, as they reliably handle a large number of edits, including \textbf{JEEP}~\cite{shi2024joint}, \textbf{EMMET}~\cite{gupta2024unified}, and \textbf{PMET}~\cite{li2023pmet}. We also include \textbf{FT-M}~\cite{zhang2024comprehensivestudyknowledgeediting} as a representative fine-tuning approach. 
While we experimented with two representative approaches—a \emph{meta-learning method} \textbf{MEND}~\cite{mitchell2022fast} and a \emph{weight-preserved method} \textbf{IKE}~\cite{zheng2023editfactualknowledgeincontext}—both achieved only 0--1\% edit success under our stricter generation-based metric, effectively amounting to complete editing failure on 1,000 edits. Consequently, we omit them from Table~\ref{tab:final result}. Details about each method, their results, and hyperparameter settings are provided in Appendix~\ref{appendix:model_not_used}.

\subsection{Main Results} 
\label{sec:results_ched_counterfact}
\paragraph{CHED and CounterFact}
Table~\ref{tab:final result} shows the results of 1,000 edits per method. When using prefix contexts composed of the exact words from the edit triplets ($s$, $o$, $o^*$), Llama3 showed declines of 6.1\% for $s$ and 38.2\% for $o$, but an improvement of 3.3\% for $o^*$, while Mistral declined by 13.3\%, 40.4\% and 5.9\%, respectively. 
Similarly, with prefix contexts consisting of hop words ($s_{hop}$, $o_{hop}$, $o^*_{hop}$), Llama3’s performance dropped by 8.42\%, 16.48\% and 2.47\%, and Mistral’s by 12.3\%, 18.2\% and 7.1\%.
While directly including $o$ causes the largest accuracy drop, the hop-word prefix contexts also significantly degrade performance. This shows that the presence of even indirectly related contexts can substantially reduce edit success.

For Llama3, CoRE achieves the highest average scores across \emph{Efficacy}, \emph{Generalization}, and \emph{Specificity} while performing competitively to MEMIT in \emph{General Ability} and \emph{Fluency}. 
While CoRE improves \emph{Efficacy} over MEMIT even when no context is prepended, the improvements are substantially greater when prefix contexts are present, suggesting its effectiveness in enhancing context robustness specifically.
EMMET shows context robust \emph{Efficacy}, but it completely breaks down for \emph{Specificity} and \emph{General Ability}.
Mistral exhibits a similar pattern, with CoRE substantially outperforming the baseline methods. While some baselines achieve better \emph{General Ability} and \emph{Fluency}, this comes at the cost of significantly reduced knowledge editing performance, which is the primary objective.

\paragraph{zsRE}  
Table~\ref{tab:zsre_results} presents the results of 1,000 edits on the zsRE dataset. Unlike CounterFact and CHED, which consist of declarative sentences, zsRE is composed of questions. As the results show, CoRE achieves the highest \emph{Efficacy}, \emph{Generalization}, and \emph{Specificity} scores. Overall, these findings further demonstrate its effectiveness in knowledge editing. More detailed results can be found in Table~\ref{tab:zsre final result} in Appendix.

\begin{table}[t]

    \centering
    \renewcommand{\arraystretch}{0.85}
    \small
    \resizebox{0.48\textwidth}{!}{%
    \begin{tabular}{l|cccc}
    \toprule

    \multirow{2}{*}{\textbf{Method}} & \multicolumn{1}{c}{\textbf{Efficacy}} & \multirow{2}{*}{\textbf{\shortstack{Generali-\\zation}}} & \multirow{2}{*}{\textbf{\shortstack{Speci-\\ficity}}} & \multirow{2}{*}{\textbf{Average}}\\ 
    & \multicolumn{1}{c}{No ctx} & \multicolumn{3}{c}{} \\

   \midrule
   
    Base & 2.7 & 3.3 & 30.3 & - \\
    \cmidrule(lr){1-5}
   MEMIT & 48.7 & 44.6 & 28.6 & \underline{40.6}\\
   JEEP & 29.9 & 19.5 & 23.8 & 24.4 \\
   PMET & 43.5 & 29.2 & \underline{29.4} & 34.0\\
   FT-M & \underline{49.5} & \underline{45.1} & 1.0 & 31.9\\
   CoRE-p+r & \textbf{50.0} & \textbf{46.0} & \textbf{30.2} & \textbf{42.1} \\
   \bottomrule

    \end{tabular}%
    }
    \caption{Performance on zsRE (Llama3).}
    \label{tab:zsre_results}
\end{table}

\subsection{User vs. Assistant Contexts} \label{sec:chat_temp}
Recent language models are typically trained for dialogues with users using \textit{instruction templates} \cite{touvron2023llama2openfoundation, grattafiori2024llama3herdmodels}.
Given that this training paradigm separates the roles of \textit{user} and \textit{assistant}, whether a prefix context is provided by the user or generated by the model might influence the model’s recall of edited knowledge.
%
For this analysis, we compare two conditions: (1) prepending a prefix context without any instruction template (original setting) and (2) presenting the context as a user utterance using the user template, followed by the assistant template for generating edited knowledge.
We use Llama-3-8B-Instruct and measure the success of knowledge editing based on the appearance of $o^*$ and the absence of $o$ within a 10-token window.

\begin{table}[t]
    \centering
    \resizebox{\columnwidth}{!}{%
      \begin{tabular}{ll|p{1cm} p{1cm} p{1.3cm}|p{1.3cm} p{1.3cm} p{1cm}}  
        \toprule
        \textbf{Method} & \textbf{Type}              & \(s_{hop}\)     & \(o_{hop}\)     & \(o_{hop}^*\)    & \(s_{hop\_chat}\) & \(o_{hop\_chat}\) & \(o_{hop\_chat}^{*}\) \\
        \midrule
        \multirow{3}{*}{\textbf{MEMIT}} &
        CHED              & 89.6 & 86.5 & 88.7 & 85.0   & 73.9   & 85.4 \\
        & Rand Hop   & 90.8 & 88.2 & 90.2 & 86.2   & 85.2   & 85.4 \\
        & Rand Cont   & 94.6 & 92.4 & 93.4 & 89.2   & 87.4   & 89.7 \\
        \midrule
        \multirow{3}{*}{\textbf{CoRE}} &
        CHED              & 95.1 & 93.8 & 96.6 & 91.2   & 84.9   & 94.6 \\
        & Rand Hop  & 94.9 & 94.5 & 96.7 & 92.6   & 90.1   & 91.9 \\
        & Rand Cont   & 96.7 & 95.4 & 96.5 & 93.4   & 92.2   & 93.5 \\
        \bottomrule
      \end{tabular}%
    }
    \caption{
    Comparison between assistant and user contexts (\S{\ref{sec:chat_temp}} \& \S{\ref{sec:word_ablation}}).
    (Rand Hop: Random hop word, Rand Cont: Random context).}
    \label{tab:results}
  \end{table}

Table~\ref{tab:results} presents the results for the original setting (subscript \textit{hop}) and the user context setting (subscript \textit{hop_chat}). The edit success rates decrease substantially for both MEMIT (row 1) and CoRE (row 4) when prefix contexts are provided in the user turn. However, CoRE narrows the performance gap compared to MEMIT, demonstrating its context robustness. 
We speculate that this phenomenon stems from language models being heavily trained to align with user preferences. As a result, they may over-attend to the same information when it is provided by the user and become more susceptible to distraction.
These findings suggest an interesting direction for future research on context robustness in chat settings. See Appendix~\ref{appendix:chat_template} for more details.

\subsection{Effects of Hop Words} \label{sec:word_ablation}
\begin{table}[t]
  \centering
  \small
    \begin{tabular}{lccc}
      \toprule
      Prefix Type      & s$_{\text{hop}}$ & o$_{\text{hop}}$ & o*$_{\text{hop}}$ \\
      \midrule
      hop-word-only    & 81.8\%           & 73.2\%           & 88.3\%           \\
      full-sentence    & 82.2\%           & 72.7\%           & 88.0\%           \\
      \bottomrule
    \end{tabular}
\caption{Comparison of edit success rates when using hop-word-only versus full-sentence prefix contexts (no-context baseline: 90.9\%) using the same editing settings as in Table~\ref{tab:hop_word_results}.}

  \label{tab:word_vs_sentence}
\end{table}
We investigate whether the decrease in \emph{Efficacy} observed when testing knowledge editing methods on CHED is merely due to the presence of prefix text or specifically influenced by the curated hop words. We conducted an ablation experiment with two settings: (1) substituting each hop word in CHED with a random word and (2) prepending random prefix contexts.

As shown in Table~\ref{tab:results}, using random words in place of the curated hop words (rows ``Rand Hop'') increases \emph{Efficacy} compared to CHED, pronounced for $o$. 
Using random contexts (rows ``Rand Cont'') further improves \emph{Efficacy}, exerting less influence on knowledge recall. 
According to these results, both hop word selection and prefix context generation are crucial in our CHED construction, with hop word selection appearing to have a more dominant effect. See Appendix~\ref{appendix:ablation} for more details.

As another experiment for examining the impact of hop words, we compared \emph{Efficacy} between prefix contexts composed solely of hop words (e.g., \emph{“Magic Johnson”}) and those using fully formed sentences generated from the same hop words (e.g., \emph{“Magic Johnson’s impact on the game \ldots”}). The results indicated that the difference in \emph{Efficacy} between these two settings is minimal, averaging around 0.4\%, suggesting that hop words alone already substantially contribute to distractiveness. Detailed numerical results can be found in Table~\ref{tab:word_vs_sentence}.

\subsection{Average Contribution Score}
\label{sec:ACS}

\begin{figure}[t]  
    \centering
    \includegraphics[width=0.48\textwidth]{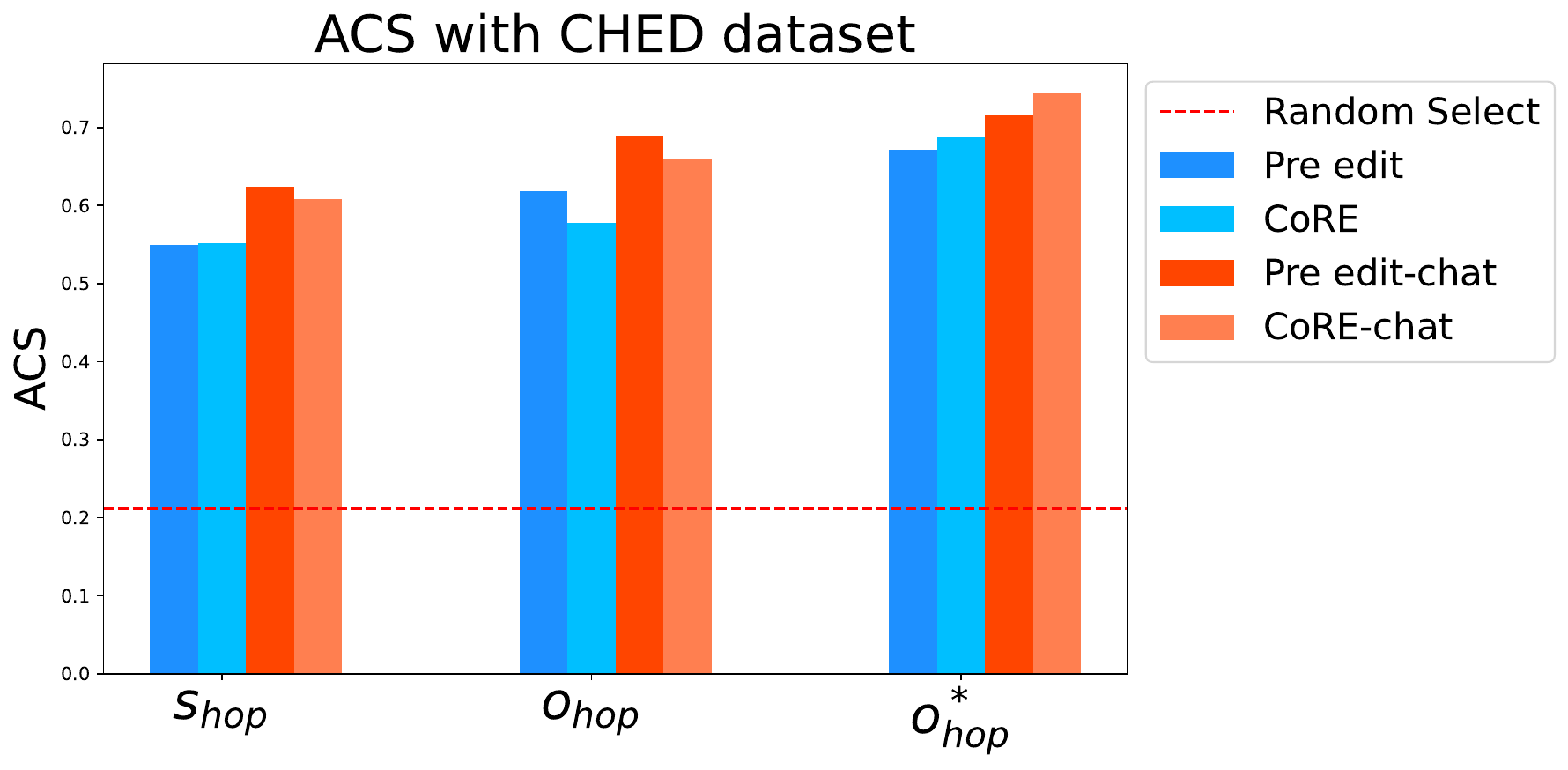} 
    \caption{ACS with CHED dataset. The dashed horizontal line represents the ACS when the model selects tokens uniformly at random within a context (0.21). }
    \label{fig:ACS}
\end{figure}

We analyze the influence of hop words more quantitatively based on attention scores. Specifically, we define a metric, Average Contribution Score (ACS), as the proportion of prefix contexts in which a hop word receives the highest attention weight among all words in the context, during the last decoding step of knowledge generation.

More specifically, we measure how strongly the final token $t_{\text{last}}$ attends to each token $t_i$ in the prefix context by aggregating $t_i$'s attention weights across all layers and heads in a pretrained Transformer model.
Formally, let $A_{\ell,h}(t_i, t_{\text{last}})$ denote the attention weight of token $t_i$ received from $t_{\text{last}}$ at layer $\ell$ and head $h$. Let $L$ be the number of layers and $H$ be the number of heads per layer. We define the token-level average attention score $\bar{A}_{i \to \text{last}}$ as:
\begin{equation}
\label{eq:avg_attn_score}
\bar{A}_{i \to \text{last}} = \frac{1}{L \cdot H} \sum_{\ell=1}^{L} \sum_{h=1}^{H} A_{\ell,h}(t_i, t_{\text{last}}).
\end{equation}

Given a knowledge editing case with a prefix context containing a hop word $t_{hop}$, the indicator $I$ of whether the hop word receives the highest attention is defined as:
\begin{equation}
\label{eq:informative_indicator}
I = 
\begin{cases}
1 & \text{if } \argmax_{i \in \text{prefix}} \bar{A}_{i \to \text{last}} = t_{\text{hop}}, \\[4pt]
0 & \text{otherwise}.
\end{cases}
\end{equation}

Finally, we define ACS as the percentage of test cases where hop words receive the highest attention:
\begin{equation}
\label{eq:acs_final}
\mathrm{ACS} = \frac{1}{N} \sum_{n=1}^{N} I_n,
\end{equation}
where $N$ is the number of test cases. This value quantifies reflects the degree of influence of hop words on knowledge recall.

In Figure~\ref{fig:ACS}, the blue bars compare the ACS of hop words before (darker) and after (lighter) knowledge editing by CoRE. Compared to random chance (red line), hop words receive significantly more attention. However, for $o_{hop}$, which is the most distractive type of hop words, the model pays less attention to them after being edited by CoRE (lighter blue), explaining CoRE's context robustness. 
Conversely, the model pays even greater attention to $o^*_{hop}$ after editing. 
Since $o^*_{hop}$ is related to the edited knowledge, it provides a useful signal for edited knowledge.
The results show that CoRE does not simply reduce the model's attention to prefix contexts; rather, it improves the model's ability to attend less to distractive information and more to useful information in the context.
The red bars in the figure represent the user context setting and show the same pattern. See Appendix~\ref{ACS_Definition} for more details.

\section{Conclusion}
We introduce and release CHED, a benchmark designed to evaluate the context robustness of knowledge editing. Our evaluation across various methods reveals that even those which perform well often fail when a prefix context is introduced. This finding underscores that the aspect measured by CHED has been largely overlooked by previous knowledge editing methods. It emphasizes the importance of this evaluation. To address this gap, we propose CoRE, which enhances context robustness. We hope that CHED, together with CoRE, will contribute to the development of more context robust, practical, and reliable knowledge editing techniques for real-world applications.

\section*{Limitations}

We built CHED using only 1-hop words extracted from Wikidata relations. 
Although any entity directly connected by a Wikidata relation is defined as a 1-hop word, this does not guarantee that the semantic relationship is strictly one hop. For example, ``U.S. First Lady'' might be linked through ``U.S. President'' to ``his spouse'', but we did not differentiate such multi-hop nuances. 
We also experimented with including 2-hop words; however, many of these words appeared only tangentially related to the corresponding entity. Consequently, it remains crucial to explore the degree and relevance of the relationship between these hop words and the edited knowledge—a promising direction for future work.
For our CoRE method, we built on the \emph{locate-then-edit} paradigm, which excels in large-scale editing while preserving overall model performance. We believe that further investigation into enhancing context robustness within other paradigms, such as \emph{meta-learning} or \emph{weight-preserving} approaches, would be a beneficial research avenue.

\section*{Ethics Statement}
Our research focuses on enhancing LLMs by rectifying errors and updating outdated knowledge through knowledge editing techniques. While these methods aim to improve user utility, they also present risks if misused, potentially generating misleading, toxic, or harmful content. It is therefore crucial to enforce strict ethical guidelines and robust safeguards to ensure that any modifications maintain overall performance and prevent the production of unsafe outputs until proper regulatory measures are established.

\section*{Acknowledgments}
This work was supported by the National Research Foundation of Korea (NRF) grant (RS-2024-00333484) and by the Institute of Information \& Communications Technology Planning \& Evaluation (IITP) grant (RS-2024-00338140, Development of Learning and Utilization Technology to Reflect Sustainability of Generative Language Models and Up-to-dateness over Time), both funded by the Korean government (MSIT).

\bibliography{main}
\nocite{*}
\appendix

\section{Details on CHED construction}

\subsection{Data Statistics} \label{appendix:data_statistics}
Table~\ref{tab:hop_word_stats} presents various statistics on the frequency distribution of the collected hop words, indicating that the distribution is highly skewed.
The skewness of the frequency in our hop words set reached 39. This indicates a highly right-skewed distribution, as skewness values greater than 1 generally suggest such behavior~\cite{groeneveld1984measuring}.

\subsection{Skewness Computation}
\label{sec:skewness}

We calculate skewness using moments to describe the shape of hop words frequency distribution. The \(k\)-th central moment of a dataset is a measure of the dataset's deviation from the mean, raised to the power \(k\). For skewness, we specifically use the third central moment and the second central moment (variance).

The data points in this context represent each word's frequency in the dataset. The number of unique words in the dataset is denoted as \( N \).

The skewness of a sample is calculated as:

\[
g_1 = \frac{m_3}{m_2^{3/2}}
\]

where:

- \( m_3 \) is the third central moment, which is calculated as:

\[
m_3 = \frac{1}{N} \sum_{n=1}^{N} (x_n - \bar{x})^3
\]

- \( m_2 \) is the second central moment, which is the variance, and is calculated as:

\[
m_2 = \frac{1}{N} \sum_{n=1}^{N} (x_n - \bar{x})^2
\]

In these formulas, \( x_n \) represents the frequency of the \(n\)-th word in the dataset, \( \bar{x} \) is the mean frequency of the words, and \( N \) is the number of unique words in the dataset. The value \( k \) refers to the order of the moment, where \( k = 2 \) corresponds to variance and \( k = 3 \) corresponds to skewness.

\begin{table}[t]
    \scriptsize 
    \begin{tabular}{c|c|c|c|c}
        \toprule
        \multicolumn{5}{c}{\textbf{Basic Word Statistics}} \\
        \midrule
        Total Words & Unique Words & Max Freq. & Min Freq. & Mean Freq. \\
        \midrule
        4,346,604 & 117,894 & 32,086 & 1 & 36.87 \\
        \midrule
        \midrule
        \multicolumn{5}{c}{\textbf{Frequency Distribution}} \\
        \midrule
        Q1 (25\%) & Median (Q2) & Q3 (75\%) & Std Dev. & Skewness \\
        \midrule
        1.0 & 1.0 & 4.0 & 289.34 & 39.29 \\
        \bottomrule
    \end{tabular}
    \caption{word set statistics.}
    \label{tab:hop_word_stats}
\end{table}

\subsection{Word Frequency}
\label{sec:word_frequency}

In the collected hop words, we observed that the most frequent terms are primarily derived from formal changes in Wikidata. For instance, the top five most frequent hop words and their respective frequencies are as follows:
\begin{itemize}
    \item ``Brockhaus and Efron Encyclopedic Dictionary'' with 32,263 times
    \item ``Small Brockhaus and Efron Encyclopedic Dictionary'' with 30,371 times
    \item ``United States of America'' with 22,407 times
    \item ``Jewish Encyclopedia of Brockhaus and Efron'' with 16,649 times
    \item ``Granat Encyclopedic Dictionary'' with 11,953 times
\end{itemize}

After excluding terms related to changes in Wikidata, the five most frequent terms are:
\begin{itemize}
    \item ``United States of America'' with 22,407 times
    \item ``United Kingdom'' with 11,150 times
    \item ``English'' with 10,664 times
    \item ``human'' with 10,096 times
    \item ``Italy'' with 9,979 times
\end{itemize}

These terms are often related to the place of birth or the native language of an entity, and therefore, they do not provide substantial contextual information about the entity.

\subsection{Frequency Test}\label{appendix:frequency}
We evaluated how the placement of high-frequency and low-frequency hop words as contextual sentences before edit sentence affects knowledge editing performance. For this experiment, we used 1,000 instances from the CounterFact dataset and applied MEMIT on Llama-3-8B-Instruct. We selected the top five most frequent and the top five least frequent hop words and constructed sentences with these words using GPT-4o mini. The evaluation measured how much the model’s ability to recall new knowledge declined when these sentences are placed before the edit prompt \((s,r)\).

Table~\ref{tab:hop_word_results} shows that when low-frequency hop words are used as prefix context, performance drops notably—especially when a sentence containing \(o_{hop}\) is placed before the edit prompt, causing edit success rate to fall to 72.7\% compared to 88.0\% when high-frequency words are used. This result supports our hypothesis that less frequent, uniquely associated hop words exert a stronger contextual influence on the model’s ability to recall edited knowledge. Based on these results, we selected hop words using frequency as the primary criterion, prioritizing those with lower occurrence counts. 

\begin{table}[t]
  \centering
  \tiny
  \resizebox{\columnwidth}{!}{%
    \begin{tabular}{lccc}
      \toprule
      Condition & s$_{hop}$ & o$_{hop}$ & o*$_{hop}$ \\
      \midrule
      Freq$\rightarrow$Sim  & 74.5& 69.1 & 78.1 \\
      Freq                 & 74.1& 69.2 & 76.9 \\
      Sim$\rightarrow$Freq  & 74.2& 69.4 & 77.0 \\
      Sim                  & 74.6 & 70.3& 77.5 \\
      Random               & 75.0& 71.7 & 77.3 \\
      Logits               & 76.2& 73.4 & 78.5 \\
      \bottomrule
    \end{tabular}%
  }
  \caption{Contextual Word Selection Methods}
  \label{tab:hop_word_selection_methods}
\end{table}

\subsection{Contextual Word Selection Methods}\label{appendix:method_ab}

Table~\ref{tab:hop_word_selection_methods} shows the edit success rates for the \(s_{hop}\), \(o_{hop}\), and \(o_{hop}^*\) sentences with our 6 word selection methods. We focus on how effectively the prefix context with contextual hop words via each selection method reduces the edit success rate after editing with \(o_{hop}\). This analysis is particularly important because the primary goal in constructing this dataset is to distract the model’s editing outcome, thereby recalling the original object. Although the Freq-Sim (69.1\%) and Freq (69.2\%) methods yield similar results on the \(o_{hop}\) sentences, the \(o_{hop}^*\) results demonstrate that the Freq-Sim method not only distracts with the \(o\) hop sentence but also with the \(o^*\) hop sentence, preventing a significant drop in the edit success rate (78.1\% for Freq-Sim versus 76.9\% for Freq). Consequently, we adopt the Freq-Sim method.

\begin{figure*}[t]  
    \centering
    \begin{minipage}{1\textwidth}
        \begin{framed}
You are tasked to create a set of sentences based on the provided **word list** that establish a natural context.

General Instructions:

1. Flow and Coherence:
    - Each sentence must smoothly lead into and set up the sentence: `\{edit_prompt\}'.
    - The generated sentences should feel like a natural precursor to the given sentence.

2. Word Usage:
    - Use each word from the **word list** exactly once, **in the exact order provided in the list**.
    - **Do not change the order** of the words in the **word list**.
    - Do not repeat any word across sentences
    - Exclude the following words entirely: `\{exclude\_words\}'.
    
3. Sentence Structure:
    - Each sentence must be concise (no longer than 20 words).
    - Avoid overly general statements or clichés (e.g., ``is known for its unique culture and history'' or ``has historical significance'').
    
4. Output:
    - Generate exactly \{len(one\_hop)\} sentences.
    - Each sentence must correspond to one word from the **word list**, in the same order as they appear in the list.
    - Return only the generated sentences, excluding the target sentence: `\{edit_prompt\}'.

Word List:
`\{one\_hop\}'

        \end{framed}
    \end{minipage}
    \caption{Prompt template for generating prefix contexts using one-hop words.}
    \label{page:prompts}
\end{figure*}

\subsection{Prefix Context Generation Methods} \label{appendix:gpt_prompt}

To construct a concise yet contextually rich sentence preceding each edit prompt, we used GPT-4o mini with the following three key constraints, providing it with a set of instructions to generate these prefix contexts systematically.  

\begin{enumerate}
    \item \textbf{Flow and Coherence}\\
    Each sentence must lead smoothly and set up the sentence: `edit prompt'.

    \item \textbf{Word Usage}\\
    The sentence must include the hop word while excluding $s$, $o$, and $o^*$.

    \item \textbf{Sentence Structure}\\
    Each sentence should be no longer than 20 words and should avoid overly general statements or clichés.
\end{enumerate}

We performed a validation process on the initially generated hop sentence dataset to ensure compliance with the Word Usage constraints. Specifically, we verified whether each hop sentence included its corresponding one-hop word while ensuring that it did not contain $s$, $o$, or $o^*$. However, if the one-hop word itself contained $s$, $o$ or $o^*$, its presence in the generated sentence was unavoidable and thus considered valid. For example, if \textit{o\(^*_{hop}\)} was ``WikiProject Football'' and $o$ was ``football'', the occurence of ``football'' in the generated sentence was permitted.

Figure~\ref{page:prompts} shows an example of the prompt we used to generate sentences with hop words. Additionally, the CHED dataset samples are shown in Figures~\ref{page:CHED_example2} and \ref{page:CHED_example3} (Our contribution is from ``sbj\_hop\_word'' to ``obj\_new\_hop\_sentence'').

\subsection{Context Coherence Evaluation}
\label{appendix:geval}

To quantitatively assess context coherence, we conducted an additional evaluation using G-Eval with GPT-4o-mini (the exact instruction prompt is provided in Figure~\ref{page:prompts2}  ). Since most knowledge facts are incomplete sentences (e.g., “Tim Cook, who works for Apple”), making coherence judgments difficult, we first generated continuations for these facts to form complete sentences using GPT-4o-mini before measuring coherence.

The coherence scores ranged from 1 (poor coherence) to 5 (excellent coherence), with the following results on a 1–5 scale:

\begin{itemize}
  \item $s$ prefix context: 4.57  
  \item $o$ prefix context: 3.85  
  \item $o*$ prefix context: 2.75  
  \item s$_{hop}$ prefix context: 3.37  
  \item o$_{hop}$ prefix context: 3.33  
  \item o*$_{hop}$ prefix context: 2.80  
\end{itemize}

The relatively low coherence scores for the edited-object contexts—both direct object-new ($o^*$) at 2.75 and hop-word–based object-new (o*$_{\!hop}$) at 2.80—are expected, because they rely on terms that are not naturally related to the original knowledge. By contrast, the original-knowledge contexts ($s$ at 4.57, $o$ at 3.85) and their hop-word variants (s$_{\!hop}$ at 3.37, o$_{\!hop}$ at 3.33) all achieve above-middle coherence. The fact that s$_{\!hop}$ and o$_{\!hop}$ do not reach even higher levels is a consequence of our using low-frequency hop words to maximize distractiveness: infrequent, highly specific terms inherently make it harder to craft fully natural sentences. Nevertheless, we prioritized distractiveness in order to rigorously evaluate context robustness, and we judge these coherence levels to be acceptable.

\begin{figure*}[t]  
    \centering
    \begin{minipage}{1\textwidth}
        \begin{framed}
"You will be given two sentences: Sentence 1 and Sentence 2. Your task is to assess 
the coherence of Sentence 2 as a continuation of Sentence 1. Evaluate how logically and 
semantically coherent Sentence 2 is with the preceding sentence using the following 
scoring system:

Evaluation Criteria:

    Coherence (1–5)
    
        1 (Incoherent)      – Sentence 2 is completely disconnected or nonsensical.
        
        2 (Barely coherent) – Sentence 2 shows minimal connection, with major shifts.
        
        3 (Moderately coherent) – Sentence 2 follows but has minor inconsistencies.
        
        4 (Largely coherent)   – Smooth continuation with only slight shifts.
        
        5 (Highly coherent)    – Perfect logical and semantic flow.

Evaluation Steps:

    1. Read both Sentence 1 and Sentence 2 carefully.
    
    2. Determine if Sentence 2 follows from Sentence 1 and maintains theme.
    
    3. Assign a score of 1–5 based on the coherence criteria above."

        \end{framed}
    \end{minipage}
    \caption{Instruction Prompt for G-Eval}
    \label{page:prompts2}
\end{figure*}

\subsection{Dataset Summary} \label{appendix:dataset_summary}
\begin{table}[t]
\centering
\resizebox{\columnwidth}{!}{%
\begin{tabular}{c|c|c|c|c|c|c}
\hline
\textbf{counts} & \textbf{1} & \textbf{2} & \textbf{3} & \textbf{4} & \textbf{5} & \textbf{Total} \\
\hline
\textit{\(s_{hop}\)} & 763 & 1,340 & 1,304 & 1,320 & 17,055 & 97,910 \\
 \(o_{hop}\) & 2 & 108 & 22 & 231 & 21,419 & 108,303 \\
    \textit{o\(^*_{hop}\)} & 2 & 129 & 35 & 273 & 21,343 & 108,172 \\

\hline
\end{tabular}%
}
\caption{CHED Dataset Size}
\label{tab:dataset_size}
\end{table}

While collecting the hop words from Wikidata, we found that some entities do not have enough full 5-hop words to form each prefix context.
In CHED, 97\% of instances have 5 prefix contexts associated with \(o_{hop}\) and \textit{o\(^*_{hop}\)}, whereas only 77\% of instances have the full set of 5 prefix contexts associated with \textit{\(s_{hop}\)}.
The relatively low number of \textit{subject hop sentences} can be attributed to the nature of factual knowledge representation—where specific words (e.g., \textit{``Danielle Darrieux''}) typically appear as subjects, whereas more general words (e.g., \textit{``English''}) function as objects—resulting in different sentence counts across categories. Consequently, for prefix contexts using hop words, we constructed a dataset of \textbf{314,385} sentences based on \textbf{21,782} fact triplets.
The details of the dataset size are provided in Table~\ref{tab:dataset_size}.

\section{Method Preliminaries}

\subsection{Full Derivation of \texorpdfstring{$k(x)$}{k(x)}}
\label{appendix:kx_derivation}

We compute \(k(x)\) as follows:
\[k(x) = \sigma\bigl(W_{fc}\,a(x) + b_{fc}\bigr),\]
\[a(x) = \gamma\Bigl(\mathrm{Att}\bigl(h^{l-1}(x)\bigr) + h^{l-1}(x)\Bigr),\]
where \(\sigma(\cdot)\) denotes a non-linear activation, and \(W_{fc}\), \(b_{fc}\) are parameters of the MLP layer. Here, \(h^{l-1}(x)\) is the hidden state at layer \(l-1\), and \(\mathrm{Att}\bigl(h^{l-1}(x)\bigr)\) is the output of the attention mechanism applied to that hidden state. We then sum the attention output with the hidden state itself and normalize via \(\gamma(\cdot)\). This process extracts the final MLP activation at the last token of the subject \(s\).

\subsection{Full KL-Divergence Term}
\label{appendix:kl_term}
Here, we expand the KL-divergence penalty \(D_{\mathrm{KL}}(\mathbf{v})\) in Equation~\eqref{eq:ori_objective_main}:
\begin{equation}
\begin{aligned}
\mathbf{v}^*
&= \argmin_{\mathbf{v}}
   \frac{1}{N} \sum_{j=1}^{N} \Bigl[
       -\log \mathbb{P}_{G(h^l = \mathbf{v})}[o^* \,\mid\, x_j + p]
   \Bigr] \\
&\quad + D_{\mathrm{KL}} \Bigl(
        \mathbb{P}_{G(h^l = \mathbf{v})}[x \,\mid\, p']
        \,\Big\|\,
        \mathbb{P}_{G(h^l)}[x \,\mid\, p']
    \Bigr),
\end{aligned}
\label{eq:ori_objective_appendix}
\end{equation}
where \(\mathbb{P}_{G(h^l=\mathbf{v})}[x \mid p']\) is the generation distribution under the modified hidden state \(\mathbf{v}\), and \(\mathbb{P}_{G(h^l)}[x \mid p']\) is the original distribution before the update. The second term minimizes the KL divergence between the output distributions for the probe prompt \(p'\) (\textit{“\{subject\} is a”}) before and after the update, thereby preventing unintended changes to related knowledge.

\section{Analysis of Prefix Context}
\label{appendix:prefix_context_c}
\subsection{Analysis of Value Vector Variance Across Different Prefix Context Strategies}
\label{appendix:extended_fig_5}
\begin{figure}[ht]
    \centering
    \includegraphics[width=0.45\textwidth]{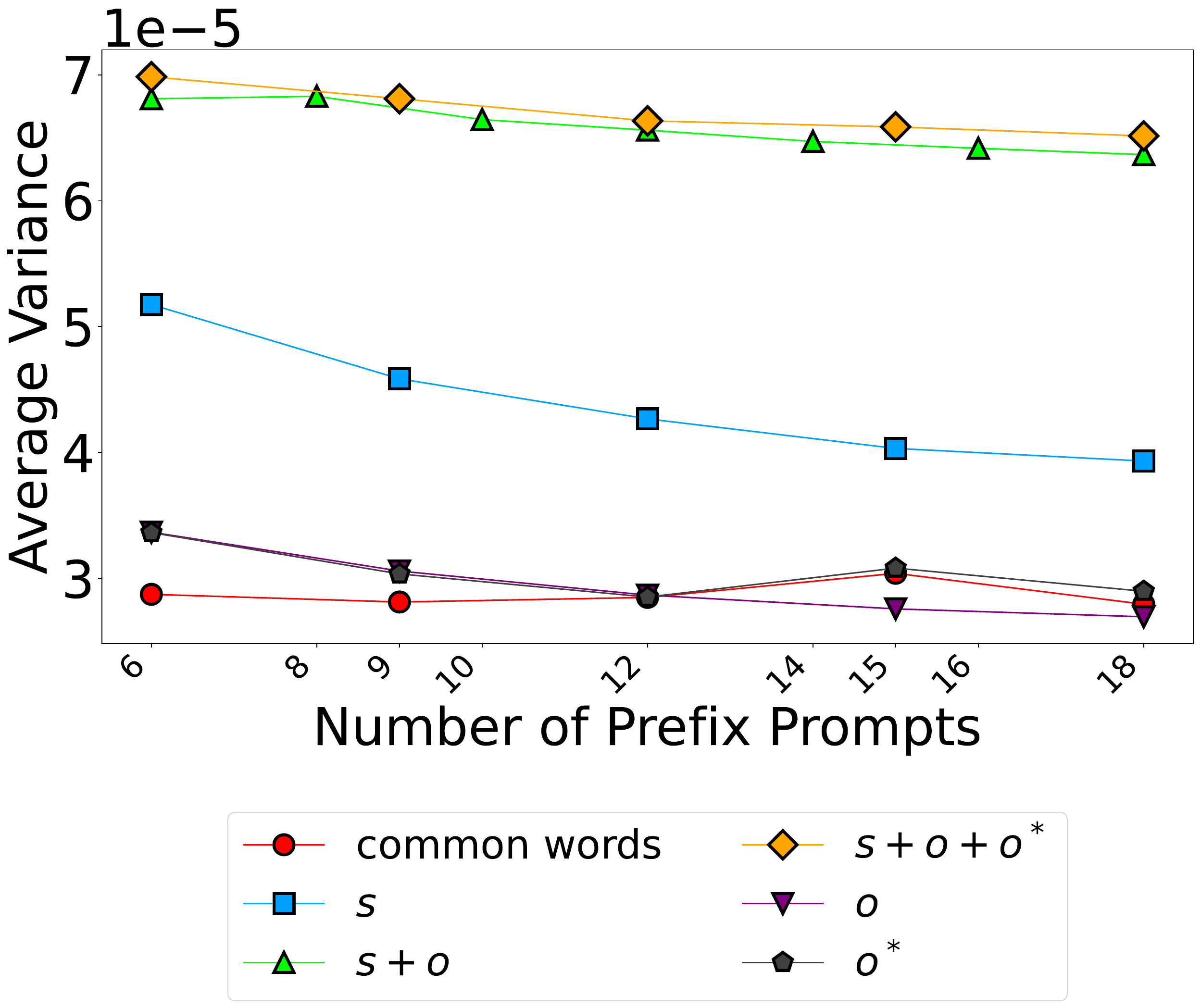} 
    \caption{Average Variance of Value Vectors by Different Prefix Prompts Strategies}
\label{fig:extended_fig_5}
\end{figure}

Figure~\ref{fig:extended_fig_5} shows an extended version of the left panel in Figure~\ref{fig:two_figures}, where the number of prefix prompts is plotted in finer detail. In this experiment, we assess whether different prefix context strategies yield greater diversity in value vectors by using 1,000 edit triplets from the CounterFact dataset. The value vectors, \(\mathbf{v}\), are extracted from the third MLP layer of Llama-3-8B-Instruct. Specifically, each strategy is constructed as follows: for the \(s\), \(o\), and \(o^*\) strategies, sentences are generated exclusively using the corresponding word. For instance, in the \(s\) strategy, all sentences are generated solely with \(s\) (e.g., producing 6 sentences using \(s\)). In contrast, the \(s, o\) strategy forms a two-sentence set—one sentence using \(s\) and one using \(o\)—while the \(s, o, o^*\) strategy forms a three-sentence set with one sentence each generated using \(s\), \(o\), and \(o^*\). In comparison, the \emph{common words} strategy from MEMIT generates sentences by selecting words from a predetermined set (e.g., ``The'', ``Therefore'', ``Because'', ``I'', ``You'').

In the combined strategies, the total number of prefix contexts increases by 2 for the \(s, o\) strategy and by 3 for the \(s, o, o^*\) strategy, starting from 6 prefix contexts for the \(s\) strategy. Notably, even when using up to 18 prefix contexts, the overall variance does not increase significantly. Since increasing the number of sentences does not significantly affect the variance, CoRE method uses 15 sentences (i.e., 5 sentences per word in the \(s\), \(o\), and \(o^*\) strategy).

\subsection{Analysis of L2 Distance among Value Vectors across Prefix Contexts}
\label{appendix:extended_fig_5_2}
\begin{figure}[ht]
    \centering
    \includegraphics[width=0.45\textwidth]{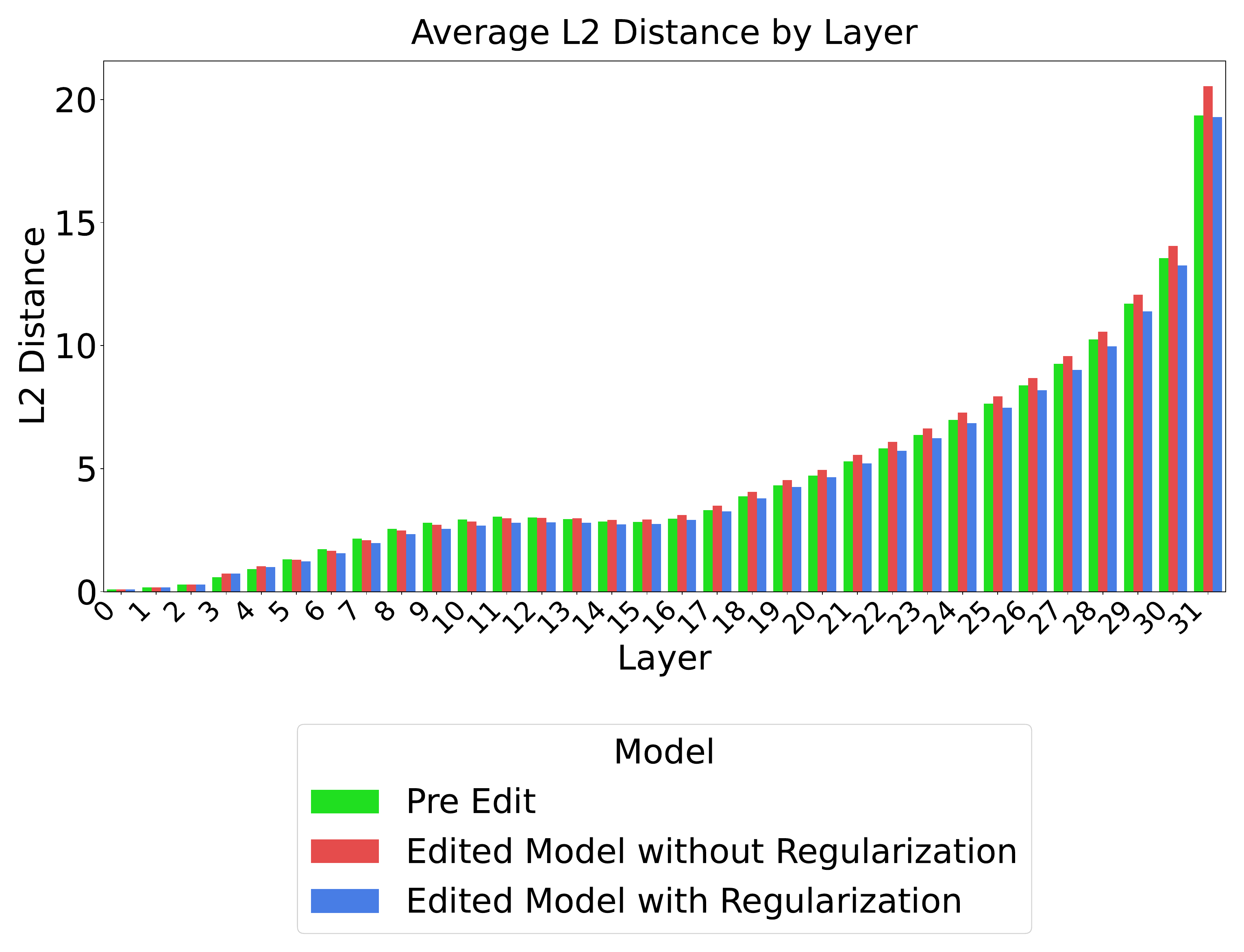} 
    \caption{Average pairwise L2 distance by layer for the pre-edit model (green), post-edit without regularization (red), and post-edit with regularization (blue).}
\label{fig:extended_fig_5_2}
\end{figure}
Figure~\ref{fig:extended_fig_5_2} shows an extended version of the right panel in Figure~\ref{fig:two_figures}. Note that the underlying experimental values remain unchanged; what differs here is the presentation. In Figure~\ref{fig:two_figures}, we focus on plotting the difference in pairwise L2 distances (post-edit versus pre-edit), whereas Figure~\ref{fig:extended_fig_5_2} presents the exact average pairwise L2 distance values for the pre-edit model, the post-edit model without regularization, and the post-edit model with our regularization term.

For further experimental details, the prefix contexts used in these experiments are from the CHED dataset, with 15 distinct prefix contexts per edit triplet.
For each input with a prepended prefix, we extracted the hidden state corresponding to the subject's last token (i.e., the value vector). Model editing was performed using MEMIT over 1,000 edits. As shown in the figure, the red bars represent the post-edit model without regularization, while the blue bars represent the post-edit model with the regularization term applied. Notably, the red bars demonstrate that, compared to the pre-edit model (represented by the green bars), the gap in L2 distances tends to increase in the later layers. In contrast, the addition of our regularization term (blue bars) effectively narrows this gap.

\section{Detailed Evaluation Metrics}
\label{appendix:eval}

In this appendix, we provide additional details and experimental setups for the evaluation metrics that are only briefly summarized in ~\S{\ref{sec:metrics}}. 

\begin{table}[ht]
\centering
\small
\begin{tabularx}{\linewidth}{|c|X|}
\hline
\multirow{2}{*}{\shortstack{\textbf{case\_id:} \\ 162}} 
  & \textbf{Prompt:} Guillaume Durand, who holds the position of \\ \cline{2-2}
  & \textbf{Generated Text:} pope, is a man of great authority and influence in the Catholic Church. He is the \textbf{bishop} of Rome and the spiritual leader of the worldwide Catholic Church. As such, he has the authority to make important decisions and to guide the Church in its teachings \\ \hline
\multirow{2}{*}{\shortstack{\textbf{case\_id:} \\ 311}} 
  & \textbf{Prompt:} The mother tongue of Jalil Lespert is \\ \cline{2-2}
  & \textbf{Generated Text:} Dutch. He is fluent in English and has a good understanding of \textbf{French} and German.
Jalil Lespert studied industrial engineering at the University of Antwerp (UHasselt) and earned his Ph.D. in 1993. After his \\ \hline
\multirow{2}{*}{\shortstack{\textbf{case\_id:} \\ 367}} 
  & \textbf{Prompt:} In Colombia, they understand \\ \cline{2-2}
  & \textbf{Generated Text:} English, but it's still a good idea to learn some basic Spanish phrases. The official language is \textbf{Spanish}, and it's widely spoken, especially in the cities.
In Peru, you'll find that many people speak Spanish, but you may also encounter \\ \hline
\multirow{2}{*}{\shortstack{\textbf{case\_id:} \\ 958}} 
  & \textbf{Prompt:} Greta Van Susteren works for \\ \cline{2-2}
  & \textbf{Generated Text:} BBC, becomes a British citizen
Greta Van Suster, the former BBC anchor and \textbf{CNN} host, has become a British citizen. She has been working for the BBC since 2015 and has been a British resident since then.
Suster, \\ \hline
\multirow{2}{*}{\shortstack{\textbf{case\_id:} \\ 13839}} 
  & \textbf{Prompt:} Melilla is a part of the continent of \\ \cline{2-2}
  & \textbf{Generated Text:} Europe, and is located in the northernmost part of Africa. It is a Spanish autonomous city located on the Mediterranean coast, and is the northernmost city in \textbf{Africa}. Melilla has a population of around 80,000 people, and is known \\ \hline
\end{tabularx}
\caption{Example cases illustrating the evaluation approach with N-gram Repetition scores.}
\label{tab:example_of_token_error}
\end{table}

\subsection{Efficacy}
As we mentioned, we propose a strict generation-based evaluation: an edit is considered successful only if the model’s output (up to 50 tokens) includes \(o^*\) while completely excluding \(o\). This approach prevents cases where the model initially produces \(o^*\) but later reverts to \(o\), as detailed in Table~\ref{tab:example_of_token_error}. 

Under this evaluation method, we evaluate edited knowledge using exact edit prompts (e.g. \textit{``Tim Cook, who works for''}) in four different conditions. The baseline condition, \emph{No ctx}, uses only the edit prompt without any additional context. The other three conditions—\(s_{hop}\), \(o_{hop}\), and \(o^*_{hop}\)—prepend different prefix contexts from our CHED dataset.
 
\subsection{Generalization}
Generalization extends the Efficacy metric by evaluating whether the model produces \(o^*\) when the edit prompt is paraphrased. For example, consider the paraphrased prompt \textit{``Tim Cook, who is employed by''} as a variant of the original edit prompt.

\subsection{Specificity} Specificity measures whether the knowledge that should remain unchanged is still the same after the edit, which is verified by asking about another subject that shares the same relation and object as in the edit prompt. For example, if the edit prompt involves a relation like “works for” with a particular object, we might ask about \textit{``Kevan Parekh, who works for''}.

\subsection{General Ability}
To verify the model’s fundamental capabilities after editing, we evaluate its performance across five key areas: commonsense reasoning, factual knowledge retrieval, context handling ability, multitask capabilities of language models across diverse subjects, and code generation. Specifically, we use CommonsenseQA \cite{talmor-etal-2019-commonsenseqa} for commonsense reasoning and TriviaQA \cite{joshi-etal-2017-triviaqa} for factual recall. We further assess long-context handling ability on the LAMBADA \cite{paperno2016lambadadatasetwordprediction}, an open-ended cloze task requiring prediction of a held-out word given the full passage. We evaluate multitask capabilities using the MMLU (Massive Multitask Language Understanding) benchmark \cite{hendrycks2021measuringmassivemultitasklanguage}, which measures language models’ performance across 57 diverse subjects. Finally, LiveCodeBench evaluates four aspects—code generation, self-repair, test-output prediction, and code execution—but here we assess only its code-generation component using the 880 publicly released problems. Representative examples for datasets are presented in Tables~\ref{tab:commonsenseqa}, \ref{tab:triviaqa}, \ref{tab:lambada} and  \ref{tab:livecodebench}. All evaluations were conducted using the OpenCompass \cite{2023opencompass} framework.

\subsection{Fluency}  
To ensure editing does not harm fluency, we measure the N-gram repetition as a proxy for disfluency. This metric is computed on outputs generated when the edit prompt is provided as input. Excessive repetition (see Table~\ref{tab:example_of_N-gram repetition} for examples, where cases with drastically increased N-gram repetition are highlighted) indicates degraded generation quality, so we penalize such outputs. Note that these examples were discovered when applying MEMIT to 1000 edit triplets. Furthermore, the N-gram repetition rate is computed as the sum of the repetition counts for 1-gram to 4-gram sequences.

\begin{table}[ht]
\centering
\small
\begin{tabularx}{\linewidth}{|c|X|}
\hline
\multirow{3}{*}{\shortstack{\textbf{case\_id:} \\ 465}} 
  & \textbf{Prompt:} Louis Bonaparte spoke the language \\ \cline{2-2}
  & \textbf{Generated Text:} Dutch Dutch language Dutch language Dutch language Dutch language Dutch language Dutch language Dutch language Dutch language Dutch language Dutch language Dutch language Dutch language Dutch language Dutch language Dutch language Dutch language Dutch language Dutch language Dutch language Dutch language Dutch language Dutch language Dutch \\ \cline{2-2}
  & \textbf{N-gram Repetition:} 183 \\ \hline
\multirow{3}{*}{\shortstack{\textbf{case\_id:} \\ 396}} 
  & \textbf{Prompt:} Pervez Musharraf follows the religion of \\ \cline{2-2}
  & \textbf{Generated Text:} Buddhism and has been the Ambassador of India to China. He has been the Ambassador of India to China. He has been the Ambassador of India to China. He has been the Ambassador of India to China. He has been the Ambassador of India to China. \\ \cline{2-2}
  & \textbf{N-gram Repetition:} 130 \\ \hline
\multirow{3}{*}{\shortstack{\textbf{case\_id:} \\ 888}} 
  & \textbf{Prompt:} The capital of Second Spanish Republic is \\ \cline{2-2}
  & \textbf{Generated Text:} Paris. The capital of the French Republic is Paris. The capital of the Kingdom of France was Paris. The capital of the Kingdom of France was Paris. The capital of the Kingdom of France was Paris. The capital of the Kingdom of France was Paris \\ \cline{2-2}
  & \textbf{N-gram Repetition:} 117 \\ \hline
\multirow{3}{*}{\shortstack{\textbf{case\_id:} \\ 570}} 
  & \textbf{Prompt:} Hannah Point is in \\ \cline{2-2}
  & \textbf{Generated Text:} Europe. The next city is Istanbul, Turkey. The next city is Istanbul, Turkey. The next city is Istanbul, Turkey. The next city is Istanbul, Turkey. The next city is Istanbul, Turkey. \\ \cline{2-2}
  & \textbf{N-gram Repetition:} 114 \\ \hline
\multirow{3}{*}{\shortstack{\textbf{case\_id:} \\ 329}} 
  & \textbf{Prompt:} Angela Merkel was employed in \\ \cline{2-2}
  & \textbf{Generated Text:} Rome as the Prime Minister of Italy, but he was not the Prime Minister of Italy. He was the Prime Minister of Italy, but he was not the Prime Minister of Italy. He was the Prime Minister of Italy, but he was not the Prime. \\ \cline{2-2}
  & \textbf{N-gram Repetition:} 113 \\ \hline
\end{tabularx}
\caption{Example cases illustrating the evaluation approach with N-gram Repetition scores.}
\label{tab:example_of_N-gram repetition}
\end{table}

\section{Analysis}

\subsection{Chat Template}
\label{appendix:chat_template}

\begin{figure}[t]
    \centering
    \begin{minipage}{0.5\textwidth}
        \fbox{%
            \begin{minipage}{\dimexpr\textwidth-2\fboxsep-2\fboxrule\relax}
edit prompt (original setting):

The new iPhone case I bought has a design that I really like. Tim Cook is employed by

Llama-3-8B-Instruct:
<|begin_of_text|>

<|start_header_id|>user<|end_header_id|>

The new iPhone case I bought has a design that I really like<|eot_id|>

<|start_header_id|>assistant<|end_header_id|>

Tim Cook is employed by

Mistral-7B-Instruct-v0.3:
<s>[INST] The new iPhone case I bought has a design that I really like[/INST]

Tim Cook is employed by
            \end{minipage}
        }
    \end{minipage}
    \caption{chat template examples}
    \label{page:Llama3 chat template}
\end{figure}

In our study, we use chat templates to investigate the impact of our dataset, with each template designed differently across various models. Figure~\ref{page:Llama3 chat template} presents an example of our chat template. During model experiments, we did not finalize the chat template to ensure that prompts aligned naturally with the model's generation process.
Also, Table~\ref{tab:results3} presents the impact of the chat template and the results of our ablation study. In Llama3, which can adapt to the chat template, we observe a decrease in the efficacy for original object sentences across all methods. This indicates that LLMs are generally influenced by the template.  
A particularly noticeable decline occurs with \(o_{hop}\) prefix context: MEMIT's success rate drops from 86.5\% to 73.9\%, and CoRE's rate decreases from 93.8\% to 84.9\%. Conversely, success rate with \(o_{hop}^*\) prefix context results in a less pronounced decrease.

\begin{table}[t]
    \centering
    \resizebox{\columnwidth}{!}{%
      \begin{tabular}{ll|cccccc}
        \toprule
        \textbf{Method} & \textbf{Type} & s & o & o* & s\_chat & o\_chat & o*\_chat \\
        \midrule
        \multicolumn{8}{c}{\textbf{Llama3}} \\
        \midrule
        \multirow{3}{*}{\textbf{MEMIT}} &
          CHED & 89.6 & 86.5 & 88.7 & 85.0 & 73.9 & 85.4 \\
          & random hop word & 90.8 & 88.2 & 90.2 & 86.2 & 85.2 & 85.4 \\
          & random sentence & 94.6 & 92.4 & 93.4 & 89.2 & 87.4 & 89.7 \\
        \midrule
        \multirow{3}{*}{\textbf{JEEP}} &
          CHED & 68.2 & 62.6 & 69.5 & 64.6 & 56.7 & 71.5 \\
          & random hop word & 63.0 & 61.3 & 60.5 & 67.0 & 60.2 & 66.0 \\
          & random sentence & 66.2 & 64.0 & 65.4 & 66.4 & 65.5 & 63.7 \\
        \midrule
        \multirow{3}{*}{\textbf{PMET}} &
          CHED & 70.2 & 66.0 & 77.3 & 64.7 & 54.2 & 75.0 \\
          & random hop word & 73.0 & 69.9 & 72.7 & 68.3 & 63.3 & 67.4 \\
          & random sentence & 74.9 & 71.8 & 72.8 & 68.7 & 66.2 & 66.5 \\
        \midrule
        \multirow{3}{*}{\textbf{EMMET}} &
          CHED & 94.2 & 93.1 & 94.6 & 94.6 & 91.1 & 96.1 \\
          & random hop word & 94.0 & 92.1 & 94.1 & 92.7 & 92.8 & 94.6 \\
          & random sentence & 93.9 & 93.1 & 93.2 & 92.5 & 93.3 & 94.2 \\
        \midrule
        \multirow{3}{*}{\textbf{CoRE}} &
          CHED & 95.1 & 93.8 & 96.6 & 91.2 & 84.9 & 94.6 \\
          & random hop word & 94.9 & 94.5 & 96.7 & 92.6 & 90.1 & 91.9 \\
          & random sentence & 96.7 & 95.4 & 96.5 & 93.4 & 92.2 & 93.5 \\
        \midrule

      \end{tabular}%
    }\
    \caption{Efficacy with chat template and hop word ablation}  
    \label{tab:results3}
  \end{table}

\subsection{Hop Words Anaylsis}
\label{appendix:ablation}

Our ablation study on hop words confirms their significant impact on the efficacy. Specifically, replacing words with hop words leads to a greater decrease in success rate compared to using random words. Moreover, the effect of hop words is comparable to replacing entire sentences at random, suggesting that the primary influence of the CHED dataset stems from the hop words themselves.

As we expected, the main contribution of our CHED dataset comes from the contextual hop-word. If we look at the \(o_{hop\_chat}\) column of Table~\ref{tab:results3}, we observe the most significant difference in \(o_{hop}\) and \(o_{hop\_chat}\) contexts, particularly when used with chat templates. In the CHED dataset, the MEMIT method shows an increase in success rate from 73.9\% to 85.2\% when using a random hop word, which is close to the 87.4\% observed in fully random contexts. Similarly, the CoRE method follows the same pattern, increasing from 84.9\% to 90.1\% with a random hop word, which is comparable to the 92.2\% achieved with fully random contexts. These results suggest that hop words act as key elements that distract the model’s attention, leading it to recall the original object despite the applied knowledge edit.

\begin{table}[t]
    \centering
    \resizebox{\columnwidth}{!}{%
      \begin{tabular}{ll|cccccc}
        \toprule
        \textbf{Method} & \textbf{Type} & s & o & o* & s\_chat & o\_chat & o*\_chat \\
        \midrule
        \multicolumn{7}{c}{\textbf{Llama3}} \\
        \midrule
        \multirow{3}{*}{\textbf{MEMIT}} &
        CHED & 76.3 & 70.6 & 72.0 & 71.4 & 63.9 & 68.4 \\
        & random hop word & 73.1 & 70.0 & 71.5 & 70.8 & 67.1 & 68.2 \\
        & random sentence & 80.0 & 78.0 & 80.7 & 73.0 & 71.8 & 72.9 \\
        \midrule
        \multirow{3}{*}{\textbf{JEEP}} &
        CHED & 55.3 & 50.0 & 54.7 & 54.6 & 48.6 & 56.8 \\
        & random hop word & 48.5 & 44.9 & 46.2 & 53.0 & 48.3 & 51.0 \\
        & random sentence & 53.7 & 51.5 & 51.7 & 53.3 & 52.2 & 51.6 \\
        \midrule
        \multirow{3}{*}{\textbf{PMET}} &
        CHED & 64.6 & 58.7 & 66.4 & 56.2 & 49.1 & 58.5 \\
        & random hop word & 63.1 & 59.4 & 62.2 & 58.3 & 52.3 & 54.6 \\
        & random sentence & 66.7 & 62.9 & 65.1 & 57.9 & 54.9 & 55.3 \\
        \midrule
        \multirow{3}{*}{\textbf{EMMET}} &
        CHED & 82.2 & 79.6 & 80.7 & 86.6 & 83.9 & 85.8 \\
        & random hop word & 80.0 & 77.5 & 79.1 & 84.7 & 82.2 & 83.9 \\
        & random sentence & 80.9 & 79.5 & 79.8 & 84.7 & 84.2 & 84.8 \\
        \midrule
        \multirow{3}{*}{\textbf{CoRE}} &
        CHED & 92.0 & 89.7 & 90.5 & 86.8 & 81.7 & 84.2 \\
        & random hop word & 90.8 & 88.7 & 90.1 & 85.4 & 82.4 & 83.0 \\
        & random sentence & 93.1 & 92.1 & 92.3 & 86.8 & 85.3 & 86.3 \\
        \midrule
      \end{tabular}%
    }
    \caption{Average probability of various methods}
    \label{tab:results_prob}
  \end{table}

\subsection{Probability Test}

In some studies, the outcome of knowledge editing is also evaluated by examining the probability difference between the original and new object tokens, thereby capturing the intrinsic differences between the two objects that are not simply generated by the language models. Accordingly, we conducted several experiments to assess not only the efficacy but also the probability of the new object token for Llama3. Especially, our method CoRE almost outperforms other methods, except for EMMET, which has a lower generalization score in the experiments. The results are presented in Table~\ref{tab:results_prob}.

\subsection{ACS} \label{ACS_Definition}

Recent studies suggest that \textit{information flow}, particularly the attention from the \textit{subject token} to the \textit{last token} of the sentence, plays a crucial role in LLM's generative performance~\cite{geva2023dissecting}. Based on this, we further investigated the influence of hop words on knowledge editing performance by measuring the Average Contribution Score (ACS). If a hop word spans multiple tokens, we compute its total impact by summing the contributions of each constituent token.





\begin{table}[htbp]
  \centering
  \adjustbox{max width=0.5\textwidth}{%
    \begin{tabular}{lcccccccc}
      \toprule
      Method & s & s\_chat & o & o\_chat & o* & o*\_chat & random & random\_chat \\
      \midrule
      \multicolumn{9}{c}{\textbf{Llama3}} \\
      \midrule
      No edit & 0.549 & 0.624 & 0.618 & 0.689 & 0.671 & 0.715  & 0.406 & 0.537 \\
      JEEP & 0.662 & 0.692 & 0.727 & 0.776 & 0.738 & 0.775  & 0.563 & 0.650 \\
      PMET & 0.602 & 0.654 & 0.698 & 0.745 & 0.736 & 0.764  & 0.488 & 0.602 \\
      MEMIT & 0.549 & 0.647 & 0.591 & 0.697 & 0.685 & 0.751  & 0.447 & 0.608 \\
      EMMET & 0.544 & 0.553 & 0.543 & 0.580 & 0.587 & 0.638  & 0.389 & 0.474 \\
      CoRE & 0.552 & 0.608 & 0.578 & 0.659 & 0.688 & 0.745  & 0.466 & 0.580 \\
      \midrule
      \multicolumn{9}{c}{\textbf{gpt-j}} \\
      \midrule
      No edit & 0.514 & - & 0.616 & - & 0.684 & -  & 0.491 & - \\
      JEEP & 0.438 & - & 0.498 & - & 0.584 & -  & 0.345 & - \\
      PMET & 0.474 & - & 0.529 & - & 0.617 & -  & 0.379 & - \\
      MEMIT & 0.460 & - & 0.584 & - & 0.662 & -  & 0.422 & - \\
      EMMET & 0.439 & - & 0.525 & - & 0.637 & -  & 0.383 & - \\
      CoRE & 0.430 & - & 0.529 & - & 0.635 & -  & 0.394 & - \\
      \bottomrule
    \end{tabular}
  }
  \caption{ACS of the various model and methods}
  \label{tab:ACS_score_results}
\end{table}

Table~\ref{tab:ACS_score_results} presents the total ACS of Llama3 and GPT-J, using various editing methods. In our CHED dataset, the average sentence length is 14.39 tokens, while the average length of hop words is 3.04 tokens. This means that when the model attends to every token randomly, the ACS with random tokens is about 0.21. As discussed in section~\S{\ref{sec:ACS}}, our model achieves a decrease in the original object's ACS and an increase in the new object's ACS in both the no-template and chat-template settings. In contrast, other methods generally exhibit either a decrease in both or an increase in both.

Notably, the CoRE method uniquely demonstrates this tendency in both simple prefix and user utterance contexts, whereas other methods achieve ACS values that are either too high, meaning they pay excessive attention to outdated $o_{hop}$ information, or too low, indicating that they disregard the $o_{hop}^*$ information after knowledge editing.

We also observe that all ACS values are higher when the prefix context is prepended as a user utterance. This indicates that the model pays more attention to the hop word, which comes from the user, suggesting that large language models extract more information from user-provided texts. Additionally, we can observe that the model achieved an increasing ACS for $o_{hop}^*$ and a decreasing ACS for $o_{hop}$ after editing, which further validates our expectations.

We speculate that this result was achieved because our CoRE method uses multiple context sentences to guide the model on which token of the context it should focus for the newly edited knowledge.

For GPT-J, we did not observe a significant difference in model behavior after editing, as the ACS decreased across all methods. We speculate that this phenomenon occurs because GPT-J is less powerful than Llama3, making it less robust to model editing. As a result, it loses its internal generality after editing.

From this, we can conclude that the attention score can be used to diagnose a model's differences after applying editing methods—not only in terms of probability or generation efficacy but also in understanding the model's internal mechanisms.

\begin{table*}[!t]
    \renewcommand{\arraystretch}{0.85}
    \small
    \resizebox{\textwidth}{!}{%
    \begin{tabular}{ll|ccccccc|cc|cc|c}
    \toprule
   
    \multirow{2}{*}{\textbf{\shortstack{Base\\Model}}} & \multirow{2}{*}{\textbf{Method}} & \multicolumn{7}{c|}{\textbf{Efficacy}} & \multirow{2}{*}{\textbf{\shortstack{Generali-\\zation}}} & \multirow{2}{*}{\textbf{\shortstack{Speci-\\ficity}}} & \multicolumn{2}{c|}{\textbf{General Ability}} & \multirow{2}{*}{\textbf{Fluency}} \\
    \cmidrule(lr){3-9} \cmidrule(lr){12-13}
    
    & & \multicolumn{1}{c}{No ctx} & \multicolumn{1}{c}{s} & \multicolumn{1}{c}{o} & \multicolumn{1}{c}{o*} & \multicolumn{1}{c}{$s_{hop}$} & \multicolumn{1}{c}{$o_{hop}$} & \multicolumn{1}{c|}{$o_{hop}^*$} & & & \multicolumn{1}{c}{C-QA} & \multicolumn{1}{c|}{T-QA} & \multicolumn{1}{c}{N-gram}\\
    
   \midrule
   
    \multirow{8}{*}{\textbf{GPT-J}}
    & Base & 0.9 & 1.6 & 0.54 & 38.06 & 1.2 & 1.0 & 10.6 & 1.1 & 26.1 & 21.5 & 32.7 & 7.6 \\
    \cmidrule(lr){2-14}
    & MEMIT & 92.8 & 77.26 & 48.9 & 85.8 & 75.5 & 69.3 & 81.4 & 64.2 & 26.3 & \underline{21.9} & 31.9 & 7.3 \\
    & JEEP & 84.9 & 75.64 & 54.88 & 84.7 & 74.7 & 70.5 & 82.3 & 63.9 & \textbf{27.2} & 20.8 & 31.0 & \textbf{7.1} \\
    & PMET & 90.4 & 79.84 & \underline{59.54} & \underline{88.82} & 81.2 & 76.6 & \underline{86.9} & \underline{70.4} & \underline{26.8} & 20.0 & 31.9 & 7.3 \\
    & EMMET & \textbf{95.3} & \textbf{81.4} & \textbf{61.14} & \textbf{91.1} & \textbf{83.6} & \textbf{79.2} & \textbf{89.1} & \textbf{73.5} & 21.8 & 19.9 & 29.7 & 7.2 \\
    & FT-M & 32.9 & 28.6 & 26.44 & 24.28 & 26.4 & 24.2 & 23.3 & 17.0 & 12.3 & 19.2 & 5.9 & 60.5 \\
    & CoRE-p & \underline{94.3} & 79.32 & 54.31 & 88.49 & \underline{81.6} & 76.0 & 85.1 & 66.3 & 24.7 & \textbf{22.0} & \textbf{32.0} & 7.2 \\
    & CoRE-p+r & 93.8 & \underline{80.68} & 58.96 & 89.54 & 81.5 & \underline{76.7} & 85.2 & 68.7 & 24.8 & \underline{21.9} & \textbf{32.0} & \textbf{7.1} \\
    \bottomrule
    \end{tabular}%
    }
    \caption{Results on GPT-J}
    \label{tab:gptj_result}
\end{table*}

\label{appendix:zsre_result}
\begin{table}[t]
    \centering  
    \renewcommand{\arraystretch}{0.4}  
    \setlength{\tabcolsep}{2pt}  
    \tiny 
    \resizebox{0.45\textwidth}{!}{
    \begin{tabular}{ll|cccc|c}
    \toprule
    
    \multirow{2}{*}{\textbf{Model}} & \multirow{2}{*}{\textbf{Method}} 
    & \multicolumn{1}{c}{\textbf{Efficacy}} & \multirow{2}{*}{\textbf{\shortstack{Gen}}} & \multirow{2}{*}{\textbf{\shortstack{Spe}}} & \multirow{2}{*}{\textbf{Avg}} 
    & \multicolumn{1}{c}{\textbf{Fluency}} \\

    & & \multicolumn{1}{c}{No ctx} & & & & \multicolumn{1}{c}{N-gram}\\

    \midrule
    
    \multirow{6}{*}{\textbf{Llama3}}
    & Base &  2.7 & 3.3 & 30.3 & - & 15.4 \\
    \cmidrule(lr){2-7}
    & MEMIT & 48.7 & 44.6 & 28.6 & \underline{40.6} & \underline{26.6} \\
    & JEEP & 29.9 & 19.5 & 23.8 & 24.4 & 27.2 \\
    & PMET & 43.5 & 29.2 & \underline{29.4} &  34.0 & \textbf{24.7} \\
    & FT-M & \underline{49.5} & \underline{45.1} & 1.0 & 31.9 & 78.8 \\
    & CoRE & \textbf{50.0} & \textbf{46.0} & \textbf{30.2} & \textbf{42.1} & \underline{26.6} \\
    \midrule\midrule

    \multirow{6}{*}{\textbf{Mistral}} 
    & Base & 1.4 & 2.0 & 23.0 & - & 4.8 \\
    \cmidrule(lr){2-7}
    & MEMIT & 40.2 & \underline{35.4} & \underline{20.8} & \underline{32.2} & \textbf{5.4} \\
    & JEEP & 20.8 & 14.4 & 20.6 & 18.6 & 6.7 \\
    & PMET & \underline{41.2} & 28.9 & \textbf{23.3} & 31.1 & 5.8 \\
    & FT-M & \textbf{48.9} & \textbf{38.2} & 10.4 & \textbf{32.5} & 16.6 \\
    & CoRE & 40.5 & \underline{35.4} & 19.9 & 32.0 & \textbf{5.4} \\
    \midrule\midrule

    \multirow{6}{*}{\textbf{GPT-J}}
    & Base & 1.2 & 0.7 & 2.9 & - & 7.6 \\
    \cmidrule(lr){2-7}
    & MEMIT & 55.0 & \underline{37.4} & \underline{3.5} & 32.0 & \underline{8.8} \\
    & JEEP & \textbf{66.8} & 36.5 & 3.4 & \textbf{35.6} & \textbf{8.3} \\
    & PMET & \underline{60.8} & \textbf{37.8} & 2.8 & \underline{33.8} & 9.6 \\
    & EMMET & 59.6 & 31.2 & 2.3 & 31.0 & 8.9 \\
    & FT-M & 15.6 & 11.2 & 1.8 & 9.5 & 41.6 \\
    & CoRE & 53.1 & 36.7 & \textbf{3.8} & 31.2 & 9.4 \\
    \bottomrule
\end{tabular}%
}
\caption{Results on zsRE}
\label{tab:zsre final result}
\end{table}

\section{Implementation Details} \label{appendix:implementation_details}
All experiments are conducted on NVIDIA A100 GPUs. Model inference was performed using vLLM \cite{kwon2023efficient}, while the probabilistic experiments were carried out using HuggingFace.

\label{sec:appendix:implementation_details}

\subsection{Mass-Editing Memory In a Transformer (MEMIT)}
\label{sec:memit}
On Llama3 and Mistral, MEMIT hyperparameters follow those used for Llama2-7b in the EasyEdit open source code~\cite{wang2024easyediteasytouseknowledgeediting}, as they share similar architecture, size, and number of layers. Optimization updates are executed for 25 steps with a weight decay of \(1 \times 10^{-3}\), a KL factor of \(0.0625\), and a learning rate of \(5 \times 10^{-1}\). Training is conducted in \texttt{fp32}, while evaluation is performed in \texttt{fp16}.

Following the same EasyEdit open source code as described above, for GPT-J-6B the EasyEdit hyperparameters are configured such that optimization updates are carried out for \(25\) steps with a weight decay of \(0.5\), a KL factor of \(0.0625\), and a learning rate of \(5 \times 10^{-1}\).

We further investigated the selection of layers for editing. While earlier work~\cite{meng2022memit} employed causal tracing to pinpoint optimal layers, later studies have shown that layers identified by causal tracing do not always lead to the best editing performance~\cite{hase2023doeslocalizationinformediting}. Motivated by these findings, we revisited the layer selection process by focusing on the early-to-mid layers. Building on prior work~\cite{gupta2024unified, yoon2024biggereditbatchsize}, we experimented with subsets consisting of 1, 2, 3, or 4 layers. For each subset, we evaluated performance based on three normalized metrics—Efficacy (no-context), General Ability, and N-gram Repetition—and computed an average score. This evaluation led us to select the following layers for editing: MEMIT: $[3]$, Mistral-7b: $[4,5]$, and GPT-J: $[2,3,4]$.

\subsection{Context Robust Editing (CoRE)}

For fairness, we use the same hyperparameters as those employed in MEMIT (\textit{see~\S{\ref{sec:memit}}}). Our method builds on these settings by incorporating an additional regularization term. In this term, the layer range and the scaling factor—denoted as \(\mathcal{L}\) and \(\lambda\) respectively in Equation~\ref{eq:prefix_loss}—were determined via parameter search using the same approach as that employed for layer selection in MEMIT.

In our experiments, we explored three configurations for the layer range: the 10 layers immediately following the edited layer, the 20 layers immediately following it, and all layers until the end of the model. Specifically, for Llama3, the chosen configuration was the 28 layers following the edited layer (layer 3) with a scaling factor of 0.04. For Mistral, the layer range comprised the 26 layers after the last edited layer (layer 5) with a scaling factor of 0.1. For GPT-J, the layer range consisted of the 26 layers following the last edited layer (layer 4) with a scaling factor of 0.0002.
The scaling factor was initially explored from 1, decrementing by 0.1. For GPT-J, since no suitable parameter was found in the initially explored range, we further refined the search starting from 0.1 in decrements of 0.01, and then from 0.01 in decrements of 0.0001. We observed a consistent trend: as the scaling factor increased, the editing success in the no-context setting tended to decrease, while metrics such as General Ability and N-gram Repetition improved.

\subsection{Equality-contrained Mass Model Editing algorithm for Transformers (EMMET)}
On Llama3 and Mistral, EMMET hyperparameters follow those used for Llama2-7b in the EMMET open source code~\cite{gupta2024unified}, as they share the similar architecture, size, and number of layers. Updates are executed at layer 5, where optimization proceeds for 25 steps with a weight decay of $1 \times 10^{-3}$, KL factor of $0.0625$, and learning rate of $5 \times 10^{-1}$. EMMET applies an emmet lambda of $0.1$. Training is conducted in \texttt{fp32}, while evaluation is performed in \texttt{fp16}.

Following the same EMMET open source code as described above, for  the EMMET hyperparameters are configured such that updates are executed at layer \(5\). Optimization is carried out for \(25\) steps with a weight decay of \(0.5\), a KL factor of \(0.0625\), and a learning rate of \(5 \times 10^{-1}\). Additionally, an emmet lambda of \(0.1\) is applied.

.

\subsection{Joint knowledge editing for information Enrichment and probability Promotion (JEEP)}
JEEP hyperparameters follow those used for Llama2-7b in the JEEP open source code~\cite{shi2024joint}, as Llama3 and Mistral share the similar architecture, size, and number of layers. Updates are executed at layers low $[5]$ and layers high $[22,23,24]$, where optimization proceeds for 30 steps with a learning rate of $0.5$. Weight decay and KL factor are set differently for each layer range: weight decay low is $0.5$ with KL factor low of $0.0625$, while weight decay high is $0.5$ with KL factor high of $0$. Training is conducted in \texttt{fp32}, while evaluation is performed in \texttt{fp16}.

Based on the same open source code,  hyperparameters are configured to update lower layers $[3,4,5,6,7,8]$ and higher layers $[15,16]$. Optimization proceeds for $30$ steps with a weight decay of $0.5$, a KL factor of $0.0625$ for lower layers and $0$ for higher layers, and a learning rate of $5 \times 10^{-1}$. Additionally, a moment adjustment weight of $2000$ is applied across both layer ranges.

\subsection{Precise Model Editing in a Transformer (PMET)}
Similar to JEEP, PMET hyperparameters follow those used for Llama2-7b in the JEEP open source code~\cite{li2023pmet}, as LLlama3 and Mistral share the similar architecture, size, and number of layers. Updates are executed at layers $[5,6,7,8,9,10]$, where optimization proceeds for 20 steps with a weight decay of $0.5$, KL factor of $1.0$, and learning rate of $0.1$. PMET applies an NLL loss factor of $2.0$. Training is conducted in \texttt{fp32}, while evaluation is performed in \texttt{fp16}.

For , PMET hyperparameters are configured to update layers $[3,4,5,6,7,8]$, where optimization proceeds for $30$ steps with a weight decay of $0.5$, KL factor of $1.0$, and a learning rate of $2 \times 10^{-1}$. PMET applies an NLL loss factor of $1.0$. Additionally, a moment adjustment weight of $6000$ is applied. Training is conducted in \texttt{fp32}, while evaluation is performed in \texttt{fp16}.

\subsection{FT-M}
FT-M \cite{zhang2024comprehensivestudyknowledgeediting} improves upon the direct fine-tuning approach (FT-L) by training the same FFN layer, identified via causal tracing in ROME, using cross-entropy loss on the target answer with the original text masked.

FT-M hyperparameters follow those used in the EasyEdit open source code~\cite{wang2024easyediteasytouseknowledgeediting}. Training is conducted in \texttt{fp32}, while evaluation is performed in \texttt{fp16}. Updates are executed at layers [21]. Optimization updates are performed over \(25\) steps with a learning rate of $5 \times 10^{-4}$.

\subsection{Methods Excluded for Low Efficacy}
\label{appendix:model_not_used}

In our work (see Appendix~\ref{appendix:eval} and Table~\ref{tab:eval_methods}), we adopt a \textbf{generation-based} efficacy metric: an edit is deemed successful only if the model actually outputs the new object ($o^*$) and does \textbf{not} output the original object ($o$) within a 50-token window. By contrast, the MEND and IKE papers employ a \textbf{probability-based} criterion, which counts an edit as successful whenever the model assigns a higher probability to $o^*$ than to $o$, regardless of whether either string is ever generated.

Under 1,000 mass edits and using a stricter, more realistic generation-based evaluation (see Table~\ref{tab:eval_methods}), MEND achieves only \textbf{1.8\%} efficacy in the no-prefix setting, despite scoring \textbf{51.5\%} under the probability-based protocol. Similarly, IKE—a prompt-based editor—manages just \textbf{0.4\%} generation efficacy, even though it reaches \textbf{68.8\%} by probability-based scoring (Table~\ref{tab:eval_methods}). Because these near-zero generation results indicate almost complete editing failure under realistic conditions, we excluded them from Table~\ref{tab:final result}.
\begin{table}[t]
\centering
  \scriptsize
\caption{Comparison of generation-based and probability-based evaluation metrics for MEND and IKE on Llama-3-8B-Instruct.}
\label{tab:eval_methods}
\begin{tabular}{l|ccc|ccc}
\toprule
\multirow{2}{*}{\textbf{Method}} 
  & \multicolumn{3}{c|}{\textbf{Generation-based (\%)}} 
  & \multicolumn{3}{c}{\textbf{Probability-based (\%)}} \\
  & No ctx & \emph{Gen} & \emph{Spe} 
  & No ctx & \emph{Gen} & \emph{Spe} \\
\midrule
MEND & 1.8 & 0.7 & 1.3 & 51.5 & 49.3 & 50.1 \\
IKE  & 0.4 & 0.3 & 37.7 & 68.8 & 72.5 & 80.3 \\
\bottomrule
\end{tabular}
\end{table}

Similarly, we omit the EMMET method on Llama3, Mistral on the zsRE dataset. Llama3 achieves only a 0.1\% efficacy, despite scoring 57.7\% in the previous efficacy calculations. Mistral also achieves only a 0.0\% efficacy, despite scoring 49.9\% in the previous efficacy.

\begin{table*}[t]
    \centering
    \small  
    \begin{tabular}{|c|p{15cm}|}
    \hline
    \textbf{ID} & \textbf{Input Prompt \& Gold Answers} \\
    \hline
    0 & 
    \begin{minipage}[t]{\linewidth}
    \footnotesize  
    \texttt{\\[0.5pt]
    [HUMAN: ``The sanctions against the school were a punishing blow, and they seemed to what the efforts the school had made to change? A. ignore B. enforce C. authoritarian D. yell at E. avoid Answer:'']\\[0.5pt]
    BOT: ``A''\\[0.5pt]
    HUMAN: ``A revolving door is convenient for two direction travel, but it also serves as a security measure at a what? A. bank B. library C. department store D. mall E. new york Answer:''}\\[0.5pt]
    \textbf{Gold Answer:} A
    \\[0.5pt]
    \end{minipage} \\
    \hline
    1 & 
    \begin{minipage}[t]{\linewidth}
    \footnotesize
    \texttt{\\[0.5pt]
    [HUMAN: ``The sanctions against the school were a punishing blow, and they seemed to what the efforts the school had made to change? A. ignore B. enforce C. authoritarian D. yell at E. avoid Answer:'']\\[0.5pt]
    BOT: ``A''\\[0.5pt]
    HUMAN: ``What do people aim to do at work? A. complete job B. learn from each other C. kill animals D. wear hats E. talk to each other Answer:''}\\[0.5pt]
    \textbf{Gold Answer:} A
    \\[0.5pt]
    \end{minipage} \\
    \hline
    2 & 
    \begin{minipage}[t]{\linewidth}
    \footnotesize
    \texttt{\\[0.5pt]
    [HUMAN: ``The sanctions against the school were a punishing blow, and they seemed to what the efforts the school had made to change? A. ignore B. enforce C. authoritarian D. yell at E. avoid Answer:'']\\[0.5pt]
    BOT: ``A''\\[0.5pt]
    HUMAN: ``Where would you find magazines along side many other printed works? A. doctor B. bookstore C. market D. train station E. mortuary Answer:''}\\[0.5pt]
    \textbf{Gold Answer:} B
    \\[0.5pt]
    \end{minipage} \\
    \hline
    \end{tabular}
    \caption{Example of the CommonsenseQA dataset with 1-shot setting}
    \label{tab:commonsenseqa}
\end{table*}
\begin{table*}[t]
    \centering
    \small  
    \begin{tabular}{|c|p{15cm}|}
    \hline
    \textbf{ID} & \textbf{Input Prompt \& Gold Answers} \\
    \hline
    0 & 
    \begin{minipage}[t]{\linewidth}
    \texttt{\\[0.5pt]
    [HUMAN: ``Answer the question, your answer should be as simple as possible, start your answer with the prompt `The answer is '.\\
    Q: Who was the man behind The Chipmunks??'']\\[1pt]
    BOT: ``The answer is [`David Seville'].''\\[1pt]
    HUMAN: ``Answer the question, your answer should be as simple as possible, start your answer with the prompt `The answer is '.\\
    Q: Who was the man behind The Chipmunks??''}\\[1pt]
    \textbf{Gold Answers:} David Seville
    \\[0.5pt]
    \end{minipage} \\
    \hline
    1 & 
    \begin{minipage}[t]{\linewidth}
    \texttt{\\[0.5pt]
    [HUMAN: ``Answer the question, your answer should be as simple as possible, start your answer with the prompt `The answer is '.\\
    Q: Who was the man behind The Chipmunks??'']\\[1pt]
    BOT: ``The answer is [`David Seville'].''\\[1pt]
    HUMAN: ``Answer the question, your answer should be as simple as possible, start your answer with the prompt `The answer is '.\\
    Q: What star sign is Jamie Lee Curtis??''}\\[1pt]
    \textbf{Gold Answers:} Scorpio, Skorpio, Scorpio (disambiguation) \\[0.5pt]
    \end{minipage} \\
    \hline
    2 & 
    \begin{minipage}[t]{\linewidth}
    \texttt{\\[0.5pt]
    [HUMAN: ``Answer the question, your answer should be as simple as possible, start your answer with the prompt `The answer is '.\\
    Q: Who was the man behind The Chipmunks??'']\\[1pt]
    BOT: ``The answer is [`David Seville'].''\\[1pt]
    HUMAN: ``Answer the question, your answer should be as simple as possible, start your answer with the prompt `The answer is '.\\
    Q: Which Lloyd Webber musical premiered in the US on 10th December 1993??''}\\[1pt]
    \textbf{Gold Answers:} Sunset Blvd, West Sunset Boulevard, Sunset Boulevard, Sunset Bulevard, Sunset Blvd.
    \\[0.5pt]
    \end{minipage} \\
    \hline
    \end{tabular}
    \caption{Example of the TriviaQA dataset with 1-shot setting}
    \label{tab:triviaqa}
\end{table*}

\begin{table*}[t]
    \centering
    \small  
    \begin{tabular}{|c|p{15cm}|}
    \hline
    \textbf{ID} & \textbf{Input Prompt \& Gold Answers} \\
    \hline
    0 & 
    \begin{minipage}[t]{\linewidth}
    \texttt{Please complete the following sentence:}\\
    \texttt{In my palm is a clear stone, and inside it is a small ivory statuette. A guardian angel.}\\
    \texttt{"Figured if you're going to be out at night getting hit by cars, you might as well have some backup."}\\
    \texttt{I look at him, feeling stunned. Like this is some sort of sign. But as I stare at Harlin, his mouth curved in a confident grin, I don't care about}\\[4pt]
    \textbf{Gold Answer:} signs \\ 
    \end{minipage} \\
    \hline
    1 & 
    \begin{minipage}[t]{\linewidth}
    \texttt{Please complete the following sentence:}\\
    \texttt{Give me a minute to change and I'll meet you at the docks." She'd forced those words through her teeth.}\\
    \texttt{"No need to change. We won't be that long."}\\
    \texttt{Shane gripped her arm and started leading her to the dock.}\\
    \texttt{"I can make it there on my own,}\\[4pt]
    \textbf{Gold Answer:} Shane \\ 
    \end{minipage} \\
    \hline
    2 & 
    \begin{minipage}[t]{\linewidth}
    \texttt{Please complete the following sentence:}\\
    \texttt{"Only one source I know of that would be likely to cough up enough money to finance a phony sleep research facility and pay people big bucks to solve crimes in their dreams," Farrell concluded dryly.}\\
    \texttt{"What can I say?" Ellis unfolded his arms and widened his hands. "Your tax dollars at work."}\\
    \texttt{Before Farrell could respond, Leila's voice rose from inside the house.}\\
    \texttt{"No insurance?" she wailed. "What do you mean you don't have any}\\[4pt]
    \textbf{Gold Answer:} insurance \\
    \end{minipage} \\
    \hline
    \end{tabular}
    \caption{Example of the LAMBADA dataset}
    \label{tab:lambada}
\end{table*}

\begin{table*}[t]
    \centering
    \small  
    \begin{tabular}{|c|p{15cm}|}
    \hline
    \textbf{ID} & \textbf{Input Prompt} \\
    \hline
    0 & 
    \begin{minipage}[t]{\linewidth}
    \texttt{\#\#\# Question: You are given a binary string s of length n and an integer numOps.}\\
    \texttt{You are allowed to perform the following operation on s at most numOps times:}\\
    \texttt{Select any index i (where 0 <= i < n) and flip s[i]. If s[i] == '1', change s[i] to '0' and vice versa.}\\
    \texttt{You need to minimize the length of the longest substring of s such that all the characters in the substring are identical.}\\
    \texttt{Return the minimum length after the operations.}\\
    \texttt{Example 1:}\\    
    \texttt{Input: s = 000001, numOps = 1}\\    
    \texttt{Output: 2}\\    
    \texttt{Explanation:}\\    
    \texttt{By changing s[2] to '1', s becomes 001001. The longest substrings with identical characters are s[0..1] and s[3..4].}\\    
    \texttt{Example 2:}\\    
    \texttt{Input: s = 0000, numOps = 2}\\    
    \texttt{Output: 1}\\    
    \texttt{Explanation:}\\    
    \texttt{By changing s[0] and s[2] to '1', s becomes 1010.}\\  
    \texttt{Example 3:}\\    
    \texttt{Input: s = 0101, numOps = 0}\\    
    \texttt{Output: 1}\\    
    \texttt{Constraints: 1 <= n == s.length <= 1000}\\    
    \texttt{s consists only of '0' and '1'.}\\  
    \texttt{0 <= numOps <= n}\\ 
    \texttt{\#\#\# Format: You will use the following starter code to write the solution to the problem and enclose your code within delimiters.}\\  
    \texttt{python}\\  
    \texttt{class Solution:}\\  
    \texttt{def minLength(self, s: str, numOps: int) -> int:}\\  
    [4pt]
    \end{minipage} \\
    \hline
    \end{tabular}
    \caption{Example of the LiveCodeBench dataset}
    \label{tab:livecodebench}
\end{table*}

\begin{figure*}[t]  
    \centering
    \begin{minipage}{1\textwidth}
        \begin{framed}

        ``case_id'': ``6'',\\
        ``counterfact_id'': ``6'',\\
        ``prompt'': ``{}, that was created in'',\\
        ``subject'': ``Anaal Nathrakh'',\\
        ``fact_knowledge'': ``Birmingham'',\\
        ``edited_knowledge'': ``Philadelphia'',\\
        ``relation_id'': ``P740'',\\
        ``rephrased_prompt'': ``In Wardha he came in  close contact with Mahatma Gandhi. Anaal Nathrakh was founded in'',\\
        ``locality_prompt'': ``City of Birmingham Symphony Orchestra, that was created in'',\\
        ``locality_ground_truth'': ``Birmingham'',\\
        ``sbj_hop_word'': [
            ``Back on Black Records'',
            ``black metal'',
            ``Season of Mist'',
            ``Candlelight Records'',
            ``United Kingdom''
        ],\\
        ``obj_old_hop_word'': [
            ``Yvonne Mosquito'',
            ``River Tame'',
            ``Changchun'',
            ``GBBHM'',
            `` West Midlands''
        ],\\
        ``obj_new_hop_word'': [
            ``Darby'',
            ``Jim Kenney'',
            ``Riverton'',
            ``USPHL'',
            ``Lower Moreland Township''
        ],\\
        ``sbj_hop_sentence'': [
            ``The label was founded to support underground artists, Back on Black Records.'',
            ``This genre is characterized by its intense sound and themes, black metal.'',
            ``The label expanded its roster significantly over the years, Season of Mist.'',
            ``Artists under this label have gained international recognition, Candlelight Records.'',
            ``The music scene in that area has a distinct identity, United Kingdom.''
        ],\\
        ``obj_old_hop_sentence'': [
            ``Yvonne Mosquito first appeared in various documentaries discussing tropical diseases.'',
            ``Residents often enjoy the beauty of the River Tame throughout the year.'',
            ``Changchun is famous for its advanced automotive industry in Asia.'',
            ``The recent events highlighted the importance of GBBHM initiatives for urban development.'',
            ``Numerous attractions can be found in the West Midlands region.''
        ],\\
        ``obj_new_hop_sentence'': [
            ``The quaint town of Darby is known for its friendly community.'',
            ``Under Mayor Jim Kenney, the city has seen significant changes.'',
            ``Located near the river, Riverton offers beautiful waterfront views.'',
            ``The USPHL provides a platform for aspiring hockey players to showcase their talent.'',
            ``Lower Moreland Township features several parks and recreational facilities.''
        ]
    
        \end{framed}
    \end{minipage}
    \caption{Example of the CHED-1}
    \label{page:CHED_example2}
\end{figure*}

\begin{figure*}[t]  
    \centering
    \begin{minipage}{1\textwidth}
        \begin{framed}

 ``case_id'': ``5644'',\\
        ``counterfact_id'': ``5698'',\\
        ``prompt'': ``{} from'',\\
        ``subject'': ``Ronan Keating'',\\
        ``fact_knowledge'': ``Australia'',\\
        ``edited_knowledge'': ``Bangladesh'',\\
        ``relation_id'': ``P495'',\\
        ``rephrased_prompt'': ``Track listing Chart References Category:2012 albums Category:Garou (singer) albums Ronan Keating was developed in'',\\
        ``locality_prompt'': ``The Slap, formulated in'',\\
        ``locality_ground_truth'': ``Australia'',\\
        ``sbj_hop_word'': [
            ``songwriter'',
            ``Boyzone'',
            ``Westlife'',
            ``voice'',
            ``singer''
        ],\\
        ``obj_old_hop_word'': [
            ``Karuwali'',
            ``Andajin'',
            ``Nyamal'',
            ``Dhungaloo'',
            ``Avstralka''
        ],\\
        ``obj_new_hop_word'': [
            ``East Bengal'',
            ``Dhaka Division'',
            ``Usui'',
            ``Oraon Sadri'',
            ``bengalese''
        ],\\
        ``sbj_hop_sentence'': [
            ``A talented songwriter crafted lyrics that resonated with many listeners.'',
            ``Boyzone became famous for their emotional ballads and captivating performances.'',
            ``Westlife captured hearts with their harmonious melodies and stunning vocal arrangements.'',
            ``Her voice captivated everyone in the studio during the recording session.'',
            ``As a singer, she expressed deep emotions through her powerful performances.''
        ],\\
        ``obj_old_hop_sentence'': [
            ``Karuwali is celebrated for its vibrant festivals held throughout the year.'',
            ``Andajin residents often gather at the marketplace to share local news.'',
            ``Nyamal stories highlight the connection between the land and its people.'',
            ``Dhungaloo offers breathtaking views that attract many nature enthusiasts each season.'',
            ``Avstralka has a diverse ecosystem that fascinates ecologists from around the world.''
        ],\\
        ``obj_new_hop_sentence'': [
            ``The history of East Bengal is rich with cultural diversity and evolution.'',
            ``Dhaka Division is known for its vibrant markets and bustling streets.'',
            ``In Japan, the art of Usui Reiki promotes healing through energy exchange.'',
            ``The Oraon Sadri community holds unique traditions that reflect their heritage.'',
            ``The Bengalese, known for their distinct language, contribute to the region's cultural tapestry.''
        ]
    
        \end{framed}
    \end{minipage}
    \caption{Example of the CHED-2}
    \label{page:CHED_example3}
\end{figure*}

\end{document}